\documentclass{UCF_ETD}
\usepackage{times}
\usepackage{graphicx}
\usepackage{subfigure}
\usepackage{multirow}
\usepackage{rotating,booktabs,multirow}

\usepackage{microtype}
\usepackage{multicol}
\usepackage{amsmath}
\usepackage{tabularx}
\usepackage{graphicx}
\usepackage{tabularx}
\usepackage{array}
\usepackage{float}
\usepackage{mathtools}
\newcommand\inner[2]{\langle #1, #2 \rangle}

\usepackage[table]{xcolor}

\newcolumntype{L}[1]{>{\raggedright\let\newline\\\arraybackslash\hspace{0pt}}m{#1}}
\newcolumntype{C}[1]{>{\centering\let\newline\\\arraybackslash\hspace{0pt}}m{#1}}
\newcolumntype{R}[1]{>{\raggedleft\let\newline\\\arraybackslash\hspace{0pt}}m{#1}}

\usepackage{cite}
\usepackage{amsmath,amssymb,amsfonts}
\usepackage{algorithmic}
\usepackage{graphicx}
\usepackage{textcomp}
\usepackage{xcolor}
\def\BibTeX{{\rm B\kern-.05em{\sc i\kern-.025em b}\kern-.08em
    T\kern-.1667em\lower.7ex\hbox{E}\kern-.125emX}}

\title{PhD dissertation, ALGORITHMS FOR INFERRING MULTIPLE MICROBIAL NETWORKS} 
\author{SAHAR TAVAKOLI} 

 \prevdegreei{ B.S. University of Tehran, 2008}
 \prevdegreeii{ M.S. Sharif University of Technology, 2014}

\thesisname{dissertation}

\degreename{Doctor of Philosophy}

\departmentsname{Computer Science}

\collegename{Engineering and Computer Science}

\termname{Summer}

\termyear{2020}

\advisorname{Shibu Yooseph}

\begin{document}

\frontmatter

\maketitle

\copyrightpage{~Sahar Tavakoli}

\begin{abstract}

\hspace{1.5 cm}The interactions among the constituent members of a microbial community
play a major role in determining the overall behavior of the community
and the abundance levels of its members. These interactions can be
modeled using a network whose nodes represent microbial taxa and edges
represent pairwise interactions.
A microbial network is a weighted graph that is constructed from a sample-taxa count matrix, and can be used to model co-occurrences and/or interactions of the constituent members of a microbial community. The nodes in this graph represent microbial taxa and the edges represent pairwise associations amongst these taxa. A microbial network is typically constructed from
a sample-taxa count matrix that is obtained by sequencing multiple biological samples and identifying taxa counts. From large-scale microbiome studies, it is evident that microbial community compositions and interactions are impacted by environmental and/or host factors. Thus, it is not
unreasonable to expect that a sample-taxa matrix generated as part of a
large study involving multiple environmental or clinical parameters can be associated with more than one microbial network. However, to our knowledge, microbial network inference methods proposed thus far assume that the sample-taxa matrix is associated with a single network. 

\hspace{1.5 cm}This dissertation addresses the scenario when the sample-taxa matrix is associated with $K$ microbial networks and considers the computational problem of inferring $K$ microbial networks from a given sample-taxa matrix. The contributions of this dissertation include 1) new frameworks to generate synthetic sample-taxa count data; 2)novel methods to combine mixture modeling with probabilistic graphical models to infer multiple interaction/association networks from microbial count data; 3) dealing with the compositionality aspect of microbial count data;4) extensive experiments on real and synthetic data; 5)new methods for model selection to infer correct value of $K$.

\end{abstract}


\begin{acknowledgments}
I would like to thank my advisor, Dr. Shibu Yooseph, for all of his guidance and support throughout the process of completing my PhD research. I would like to thank my committee members
Dr. Shaojie Zhang, Dr. Wei Zhang and Dr. George Atia for their guidance and support.
\end{acknowledgments}

\tableofcontents

\listoffigures

\listoftables

\mainmatter

\chapter{INTRODUCTION} 

\hspace{1.5 cm}Microbes are found almost everywhere on earth, including in
environments deemed too extreme for other life forms, and they play
critical roles in many biogeochemical processes \cite{falkowski2008microbial, PNAS}. Microbial communities are also found in association with higher life forms, including plants and animals;
for instance, trillions of microbes live in or on the human body
(almost as many human cells as there are in the body) \cite{Sender , sani}
and ongoing research continues to reveal the important roles that many
of these microbes play in human health \cite{huttenhower2012structure, MetaHIT, system}. Microbial
communities are typically structured and composed of members of
different species. The microbes in a community do not exist in
isolation, but interact with each other and also compete for the
available carbon and energy sources. In a microbial community, the abundance level of a constituent member is determined by its interactions with other members of the community (resulting from the competition for available resources in an environment) and/or by its interaction with the host. Revealing the nature of these interactions and co-occurrences is important for understanding the functional roles of the community members, and has implications in many areas, including in the context of human health \cite{huttenhower2012structure, sep1}. These interactions, along with
resource availability and environmental parameters (like temperature,
pH, and salinity) \cite{Hibbing, WilliamsonYooseph, meta}, determine the
taxonomic composition of the microbial community and the abundance
levels of its constituents. Knowledge of these interactions is crucial for understanding the overall behaviour of the microbial community, and can be used to elucidate the biological mechanisms underlying microbe-associated disease progression and microbe-mediated processes (like biofilm formation).

\hspace{1.5 cm}The study of microbial communities has been greatly enabled with the
advent of high-throughput next-generation DNA sequencing technologies
\cite{quail2012tale, bentley2006whole, margulies2005genome}. With advances in high-throughput DNA sequencing, it is now possible to generate large volumes of sequence data (either targeting a taxonomic marker gene \cite{woese1977phylogenetic} or whole genome shotgun sequencing \cite{venter2004environmental}) at lower costs. The taxonomic composition of a microbial community can be obtained by sequencing the DNA extracted from a biological sample that
has been collected from the environment of interest. This is
achieved either using a targeted approach, involving the sequencing of
a taxonomic marker gene (for instance, the 16S ribosomal RNA gene,
which is found in all bacteria \cite{woese1977phylogenetic, sep2}) or using a whole-genome shotgun
sequencing approach \cite{venter2004environmental}. Both approaches generate 
taxa counts that are {\em compositional} in nature, and that enable the estimation of the {\em relative} abundances of the constituent members of the community. Sequence data generated from a collection of $n$ samples defined on $d$ microbial taxa can be used to generate a sample-taxa matrix that contains counts reflecting the number of times a microbial taxon has been observed in a sample.

\hspace{1.5 cm}Microbial interactions can be modeled using a weighted graph (or
network), where each node in the graph represents a taxon (or taxonomic group) and an (undirected) edge exists between two nodes if the
corresponding taxa interact with, or influence, each other. The edge
weight captures the strength of the interaction, with its sign
reflecting whether the interaction is positive or negative. This
framework can be used to model a variety of microbial interactions, including competition and co-operation. While we do not consider it here, a directed graph can
also be used to represent interactions, where the edge direction
indicates the direction of influence (or causality). Microbial networks are typically constructed from sample-taxa count matrices. A sample-taxa count matrix is generated by sequencing
multiple biological samples ($n$ samples) collected from the environment of interest
and identifying the counts of the observed taxa ($d$ taxa) in each sample.

\hspace{1.5 cm}Microbial networks can be constructed
using a variety of different approaches. To our knowledge,
all of these methods assume that the sample-taxa matrix is associated
with a {\em single} underlying stochastic process (that is, there is
one underlying network topology and set of edge weights). However,
this need not always be the case. In this dissertation, we consider an
important extension to the network inference problem, whereby we
develop a mixture modeling framework for inferring $K$ microbial networks when the observed sample-taxa matrix is
associated with $K$ underlying distributions. Another issue that we tackle in this dissertation is finding the proper $K$(number of components) in modeling microbial count data, by introducing a new practical framework.

\hspace{1.5 cm}We are motivated by
large-scale human-associated and other environmental microbial
community projects that are now possible due to cost-effective
sequencing. For instance, human gut microbiome studies now routinely analyze large cohorts of individuals and generate microbial community data from several hundreds (to even
thousands) of samples. An important research question in this area
involves the definition of a "core" microbiome associated with a
particular host phenotype \cite{huttenhower2012structure,MetaHIT}. It is well understood that the gut
microbiome composition is greatly influenced by many factors including
diet and age, and thus it is not unreasonable to expect the
associated microbial network interactions to also be different when
these factors vary (for example, the gut microbial community
interaction network in vegetarian hosts can be expected to be different
compared to the network in non-vegetarian hosts). A similar situation also
occurs in environmental studies where the microbial interactions are
influenced greatly by the physical and chemical gradients of the
environment. Often the collected metadata in these studies may not be
comprehensive enough to discern these interactions in a supervised
manner. Our proposed mixture frameworks offer a principled approach to
identifying these multiple microbial interaction networks from a
sample-taxa matrix.

\chapter{LITERATURE REVIEW}
\hspace{1.5 cm}Sample-taxa matrices can be used to study microbial associations \cite{layeghifard2017disentangling}. These associations can be modeled using a weighted graph with $d$ nodes in which the nodes represent microbial taxa, the edges represent pairwise associations, and edge weights represent the strength of the associations. Different methods have been proposed for inferring a {\em single} microbial network from a sample-taxa matrix, including correlation based approaches \cite{gloor2017microbiome}, inference of the latent correlation structure after log-ratio transformation of the count data \cite{friedman2012inferring} \cite{fang2015cclasso}, and probabilistic graphical models  \cite{Biswas} \cite{kurtz2015sparse}.\\

\hspace{1.5 cm}Several methods have been proposed for constructing a {\em single}
microbial network from an input sample-taxa matrix \cite{Layeghifard}. One approach involves using pairwise correlations (Pearson or Spearman) between taxa to define the edge weights in the
graph. This is the most simple method that considers the pairwise correlation as similarity index to extract the interaction network. There are also other methods which calculate mutual information as similarity index or Kulback-Leiber criteria as a dissimilarity index to find the connections between the nodes in the interaction graph extracted from microbial data. However, the computation of these correlation networks directly
from the observed count data can be misleading because of the compositional nature of these data \cite{gloor2017microbiome} and causes spurious connection detection in the estimated network. Considering regular correlation as a criterion is not able to extract complex relations between the OTUs in the microbial communities. This issue is handles by multiple regression analysis methods that considers structural relation between the OTUs instead of pairwise\cite{19}. Some methods use mutual information matrices to extract a specific structure such as the Chow-Liu tree approximation \cite{chow}. This method considers higher order interactions not just pairwise which gives the opportunity for better estimation of the interaction network. However this approach forces a tree structure to the network which is not a desired case in modeling the microbial interaction networks.  \\   

\hspace{1.5 cm}Furthermore, for a microbial network with $d$ nodes, while there are $d*(d-1)/2$ edge weights that need to be determined, the number of available samples
$n$ is often not large enough, with the result that the system of
equations to determine all pairs of correlations is
under-determined. This later issue is typically handled by assuming
that the network is sparse (that is, the number of edges is $O(d)$). Methods based on latent variable modelling have been proposed
to infer correlation networks \cite{CCLASSO, SPARCC}. These methods
use log-ratio transformations of the original count data \cite{aitchison1982} to deal with
the compositional nature of these data and subsequently infer the
correlation matrix (i.e. edge weights) under the assumption of
sparsity. \\

\hspace{1.5 cm}Microbial networks have also been constructed using a
probabilistic graphical model framework \cite{Jordan} that enables the
modeling of conditional dependencies associated with the
interactions. For instance, the assumption that the log-ratio transformed
count data follow a Gaussian distribution, results in a Gaussian
graphical model (GGM) framework. In this scenario, the graph structure represents the precision matrix (or inverse covariance matrix) of the underlying multivariate Gaussian distribution. This graph has the property that an edge exists between two nodes if the corresponding entry in the precision matrix is non-zero. A zero entry in the precision matrix indicates conditional
independence between the two corresponding random variables. When the graph is assumed to be sparse, the GGM inference problem can be solved using sparse precision
matrix estimation algorithms \cite{glasso}. This approach has been used to construct microbial networks from sample-taxa matrices \cite{SPIEC-EASI} \cite{zohre1} \cite{zohre2}.\\

\hspace{1.5 cm}An alternate approach to constructing a microbial network is to model the vector of observed taxa counts (in samples) using a multivariate distribution and to infer the parameters of this
distribution from the observed data using a maximum likelihood
framework. Any candidate multivariate distribution for this approach will have to
be flexible enough to capture the underlying covariance structure to model
the network interactions (i.e., allow for both positive and negative
covariances); this rules out distributions like the multinomial or the Dirichlet-Multinomial, which are popular choices for modeling microbial count data in certain situations [\cite{DMM, Powercalc}], but which cannot capture both types of interactions. The Multivariate Poisson
Log-Normal (MPLN) distribution \cite{init} can be
used for modelling multivariate count data and its covariance structure can capture both positive and negative interactions. This distribution was used recently
\cite{biswas2015learning} to model counts in a sample-taxa matrix and infer an
underlying microbial network using an assumption of sparsity.\\

\hspace{1.5 cm}Motivated by the observation that a sample-taxa matrix generated as part of a large cohort study involving multiple environmental or clinical parameters can be associated with more than one underlying microbial network, we proposed three novel mixture model frameworks \cite{ISMB} to infer $K$ microbial networks from a sample-taxa matrix that is generated from $n$ samples and $d$ taxa. Furthermore, we introduced a new practical framework to find the best number of components and model selection in modeling microbial data with a mixture modeling framework.

\hspace{1.5 cm}In our first approach, we modeled the taxa counts in a sample using a Multivariate Poisson Log-Normal (MPLN) (described in chapter 3) distribution \cite{aitchison1982statistical, eeg}. The mixture model framework consists of $K$ MPLN distributions with different underlying parameters. We used a maximum likelihood setting to estimate the parameters, and presented an optimization algorithm based on the Minorization-Maximization (MM) principle \cite{lange2016mm}\cite{hunter2004tutorial}, and involving gradient ascent and block updates. We extend this formulation based on an $l_{1}$-penalty model and provide
algorithms to infer $K$ sparse networks. We evaluate the performance of
our algorithms using both synthetic and real datasets. We also evaluate the performance of our method
on compositional data obtained by subsampling from the true counts of
the taxa. This evaluation models the real-world scenario, where the sample-taxa matrices that we have access
to, contain only {\em relative} abundances of the observed taxa.

\hspace{1.5 cm}The second algorithm (MixMCMC) (described in chapter 4)\cite{BIBM} which is introduced in this dissertation uses Markov Chain Monte Carlo (MCMC) sampling to estimate the latent parameters in the MPLN mixture model framework, and the third algorithm (MixGGM) (described in chapter 5) \cite{BIBM} specifically addresses the compositional nature of sequencing data (i.e., the counts in a sample-taxa matrix generated by sequencing reflect {\em relative} abundances and not {\em absolute} abundances of the taxa). For this approach, the count data are first transformed by applying the Centered Log-Ratio (CLR) transformation \cite{aitchison1982statistical}. The CLR transformed matrix is assumed to follow a Multivariate Gaussian distribution, and the MixGGM algorithm uses the MM approach to estimate the parameters for a mixture of $K$ Multivariate Gaussians. In all of these methods, the underlying precision matrices of the distributions associated with the mixture components correspond to the microbial networks that we care about. As it has been noted that microbial co-occurrence or interaction networks are often sparse, we extend the MixMCMC and MixGGM approaches to enable inference of sparse networks based on an $l_1$-penalty model. Each of the three approaches (MixMPLN, MixMCMC, and MixGGM) are evaluated using synthetic datasets to assess recovery of the graphs that were used to create the data. We assess the performance of these methods on a comprehensive collection of graph types, consisting of band graphs, cluster graphs, scale-free graph and random graphs. MixMPLN and MixGGM are also applied to a real dataset. 

\hspace{1.5 cm} The last part of this dissertation (chapter 6) includes a novel model selection framework to model data with mixture of Poisson-Log Normal distribution. We have extended the Variational  Inference (VI) framework to optimise the marginal log-likelihood function and estimate the best number of components in the mixture model. 

\chapter{MM FRAMEWORK TO MODEL THE DATA WITH MIXTURE OF MPLN}

 Prior to describing our mixture modeling framework, we describe a single MPLN distribution. In our discussions, we denote matrices using upper-case letters, column vectors using bold letters (upper- or lower- case), and scalar values using normal lower-case letters.
\section{Description of the model}
{\bf Single MPLN distribution:}\\
Consider an MPLN distribution with parameter set $\Theta=(\boldsymbol{\mu},\Sigma)$, where $\boldsymbol{\mu}$ represents its $d$-dimensional mean vector and $\Sigma$ represents its $d \times d$ covariance matrix. Then, a $d$-dimensional count vector $\boldsymbol{X}=(x_{1},...,x_{d})^{T}$  generated by this distribution has the following property
\begin{equation}
\begin{split}
x_{j}|\lambda_{j} \sim \mathbb{P}(e^{\lambda_{j}})  
 \\
(\lambda_{1},...,\lambda_{d})^{T} \sim \mathbb{N}_{d}(\boldsymbol{\mu} ,\Sigma) 
\end{split}
\end{equation}
where $\mathbb{P}(c)$ denotes a Poisson distribution with mean $c$ and $\mathbb{N}_{d}(\boldsymbol{\mu} ,\Sigma)$ denotes a $d$-dimensional multivariate normal distribution with mean $\boldsymbol{\mu}$ and covariance $\Sigma$. An MPLN distribution thus has two layers, with the observed counts being generated by a mixture of independent poisson distributions whose (hidden) means follow a multivariate log normal distribution. If we use $\boldsymbol{\lambda}=(\lambda_1,\lambda_2,..\lambda_d)$ to denote the latent (or hidden) variable representing the Poisson means, then the probability density function $p(\boldsymbol{X}|\Theta)$ of the MPLN distribution is given by
\begin{equation}
 \int_{\mathbb{R}^d} \prod_{j=1}^{d}\frac{e^{-e^{\lambda _{j}}}e^{\lambda _{j}x_{j}} }{x_{j}!}
 \frac{e^{[-\frac{1}{2}(\boldsymbol\lambda-\boldsymbol\mu)^{T}\Sigma^{-1}(\boldsymbol\lambda-\boldsymbol\mu)]}}{\sqrt{(2\pi)^{d}det(\Sigma)}}d\boldsymbol\lambda
\end{equation}
Let $\boldsymbol{X_1}, \boldsymbol{X_2},..\boldsymbol{X_n}$ denote $n$ independent samples drawn from an MPLN distribution, where each $\boldsymbol{X_i}$ is a $d$-dimensional count vector.  We use $X = [\boldsymbol{X_1} \boldsymbol{X_2}..\boldsymbol{X_n}]$ to denote the sample-taxa matrix generated from $d$ taxa and $n$ samples, and $x_{ij}$ to denote the count of the $j^{th}$ taxon in $\boldsymbol{X_i}$. We can estimate the parameter set $\Theta$ of this MPLN distribution using a likelihood framework by considering the $d$-dimensional latent variables $\boldsymbol{\lambda_1},\boldsymbol{\lambda_2},..$, and $\boldsymbol{\lambda_n}$ associated with samples $\boldsymbol{X_1},\boldsymbol{X_2},..$, and $\boldsymbol{X_n}$ respectively; let matrix $\Lambda=[\boldsymbol{\lambda_1}  \boldsymbol{\lambda_2}.. \boldsymbol{\lambda_n}]$ and $\lambda_{ij}$ denote the $j^{th}$ entry in $\boldsymbol{\lambda_i}$. The log-likelihood function $L(\Theta |X,\Lambda)$ can be optimized using a simple iterative stepwise ascent (or conditional maximization) procedure to compute $\Theta$ and $\Lambda$. The estimated values of the parameters can be used to provide an approximation for $p(X|\Theta)$ as $\prod_{i=1}^{n}p(\boldsymbol{X_{i}} |\boldsymbol{\lambda_{i}},\Theta)$, where $p(\boldsymbol{X_{i}} |\boldsymbol{\lambda_{i}},\Theta)$ is defined as:
\begin{equation}
 \prod_{j=1}^{d}\frac{e^{-e^{\lambda _{ij}}}e^{\lambda _{ij}x_{ij}} }{x_{ij}!}
 \frac{e^{[-\frac{1}{2}(\boldsymbol\lambda _{i}-\boldsymbol\mu)^{T}\Sigma^{-1}(\boldsymbol\lambda _{i}-\boldsymbol\mu)]}}{\sqrt{(2\pi)^{d}det(\Sigma)}}
\end{equation}
An analogous approach based on optimizing the log-posterior using an iterative conditional modes algorithm has been proposed previously  \cite{biswas2015learning}.\\\\
\noindent{\bf Mixture of $K$ MPLN distributions (MixMPLN):}\\
In our framework, we consider a mixture model involving $K$ MPLN distributions. Let $\pi_{1},...,\pi_{K}$ represent the mixing coefficients of the $K$ components (where $\sum_{l=1}^{K}\pi _{l}= 1$), and let $p_{l}$ and $\Theta _{l}$ denote the $l^{th}$ component distribution and its parameter set. A $d$-dimensional sample vector $\boldsymbol{X}$ generated from this mixture has the following distribution:   
\begin{equation}
p(\boldsymbol{X}|\pi_{1},...,\pi_{K},\Theta_{1},...,\Theta_{K})=\sum_{l=1}^{K}\pi _{l}p_{l}(\boldsymbol{X}|\Theta _{l})
\end{equation}
For $n$ independent samples $X= [\boldsymbol{X_1} \boldsymbol{X_2}..\boldsymbol{X_n}]$, the probability distribution is given by 
\begin{equation}
p(X|\pi_{1},...,\pi_{K},\Theta_{1},...,\Theta_{K})= 
\prod_{i=1}^{n}\sum_{l=1}^{K}\pi _{l}p_{l}(\boldsymbol{X_{i}} |\Theta _{l})
\end{equation}
The general log-likelihood function is thus 
\begin{equation}
\begin{split}
L(\pi_{1},...,\pi_{K},\Lambda_{1},...,\Lambda_{K},\Theta_{1},...,\Theta_{K}|X) = \\ log(\prod_{i=1}^{n}\sum_{l=1}^{K}\pi _{l}p_{l}(\boldsymbol{X_{i}} |\boldsymbol{\lambda_{il}},\Theta _{l}))
\end{split}
\end{equation}
where $\boldsymbol{\lambda_{il}}$ is the $d$-dimensional latent variable associated with $\boldsymbol{X_i}$ in component $l$. We use a maximum log-likelihood framework to estimate the parameters of the MixMPLN model from the observed data $X$ by optimizing the function $L(\pi_{1},...,\pi_{K},\Lambda_{1},...,\Lambda_{K},\Theta_{1},...,\Theta_{K}|X)$. 
\section{Optimizing the log-likelihood function using the MM principle}
The MM principle is a general technique \cite{lange2016mm, hunter2004tutorial} that has proven to be useful for tackling function optimization problems (MM stands for Minorization-Maximization in maximization problems and  for Majorization-Minimization in minimization problems). For our scenario, let $h(\theta)$ denote a function to be maximized. The MM principle proposes to maximize the minorizer function $g(\theta|\theta^{t})$ instead of maximizing  $h(\theta)$ directly; here, $\theta^{t}$ denotes a fixed value of the parameter $\theta$. The function $g(\theta|\theta^{t})$ is said to be a minorizer of $h(\theta)$ if:
\begin{equation}
\begin{split}
    h(\theta^{t})=g(\theta^{t}|\theta^{t})\\
    h(\theta) \geq g(\theta|\theta^{t}), \theta\neq\theta^{t}
\end{split}
\end{equation}
Therefore, the first step in our MM approach is to find a minorizer function which has the required property. For this, we use the following observation that follows from the concavity property of the {\em log} function \cite{lange2016mm, MM} for $m$ non-negative values $\beta_1, \beta_2,..\beta_m$ :
\begin{equation}
 log(\sum_{i=1}^{m}\beta _{i})\geq \sum_{i=1}^{m}\frac{\beta _{i}^{t}}{\sum_{j=1}^{m}\beta_{j}^{t}}log(\frac{\sum_{j=1}^{m}\beta_{j}^{t}}{\beta_{i}^{t}}\beta_{i}) 
\end{equation}
Equation~(6) can thus be lower-bounded based on this observation:
\begin{equation}
log(\prod_{i=1}^{n}\sum_{l=1}^{K}\pi _{l}p_{l}(\boldsymbol{X_{i}} | \boldsymbol{\lambda_{il}} , \Theta _{l}))\geqslant\sum_{i=1}^{n}\sum_{l=1}^{K}w_{il}^{t}log(\frac{\pi_{l}}{w_{il}^{t}}p_{l}(\boldsymbol{X_{i}}|\boldsymbol{\lambda_{il}} , \Theta _{l}))
\end{equation}
where, weight $w_{il}^{t}$ is defined as follows:
\begin{equation}
w_{il}^t=\frac{\pi _{l}^{t} p_l(\boldsymbol{X_{i}}|\boldsymbol{\lambda_{il}^{t}},\Theta _l^t)}{\sum_{l=1}^{K}\pi _{l}^{t} p_l(\boldsymbol{X_{i}}|\boldsymbol{\lambda_{il}^{t}},\Theta _l^t)}
\end{equation}
We use the function on the right-hand side of equation 3.9 as the minorizer function for our problem. Thus, we will define the new objective function (L) to be maximized as follows:
\begin{equation} 
\begin{split}
L(\pi _{1},\pi _{2},...,\pi _{K},\Lambda_{1},...,\Lambda_{K},\Theta _{1},\Theta _{2},...,\Theta _{K})= \\
\sum_{i=1}^{n}\sum_{l=1}^{K}w_{il}^{t}log(\frac{\pi _l}{w_{il}^{t}}p_l(\boldsymbol{X_{i}}|\boldsymbol{\lambda_{il}},\Theta _l))= 
\sum_{i=1}^{n}\sum_{l=1}^{K}w_{il}^{t}(log(\frac{1}{w_{il}^{t}}))+ \\ \sum_{i=1}^{n}\sum_{l=1}^{K}w_{il}^{t}(log(\pi _l))+   \sum_{i=1}^{n}\sum_{l=1}^{K}w_{il}^{t}(log(p_l(\boldsymbol{X_{i}}|\boldsymbol{\lambda_{il}},\Theta _l)))
\end{split}
\end{equation}

\section{Steps of the MixMPLN optimization algorithm}
We used a coordinate ascent approach in conjunction with a block update strategy to optimize $L$. We present the details of parameter initialization and subsequent iterative updates below. 
 
\subsection{Parameter initialization}
The $n$ samples $\boldsymbol{X_1} \boldsymbol{X_2}..\boldsymbol{X_n}$ are partitioned into $K$ clusters (components) using the K-means algorithm. Then, the samples assigned to a component are used to estimate the parameters of that component using the moments of an MPLN distribution \cite{init}, given by the equations below; here, $\sigma_{ij}$ denotes the $ij^{th}$ entry in $\Sigma$.
\begin{equation}
\begin{split}
E(x_{i})=exp(\mu _{i}+\frac{1}{2}\sigma _{ii})=\alpha _{i}
\\
var(x_{i})=\alpha _{i}+\alpha _{i}^2\left \{ exp(\sigma _{ii}) -1 \right \}
\\
cov(x_{i},x_{j})=\alpha _{i}\alpha _{j}\left \{ exp(\sigma _{ij})-1 \right \}
\end{split}
\end{equation}

\subsection{Parameter estimation in iteration $(t+1)$}
The portion of function $L$ (in equation 11) that is dependent on the $\pi_i$'s can be optimized. 
\begin{equation}
L_{1}( \pi _{1},\pi _{2},...,\pi _{l})=\sum_{i=1}^{n}\sum_{l=1}^{K}w_{il}^{t}(log(\pi _l)))
\end{equation}
Since $\sum_l{\pi_{l}}=1$, 
we can optimize $L_1$ by introducing a Lagrange multiplier and identifying a stationary point of the subsequent Lagrangian \cite{lagrang}. This yields
\begin{equation}
\pi _{l}^{t+1}=\frac{1}{n}\sum_{i=1}^{n}w_{il}^{t}
\end{equation}
where $w_{il}^{t}$ is calculated using equation~(10).\\\\
Considering the part of the L function that is dependent on $\Lambda$ and $\Theta$, we have the following term to maximize:
\begin{equation}
L_{2}(\Lambda_{1},...,\Lambda_{K},\Theta_{1},...,\Theta_{K})=\sum_{i=1}^{n}\sum_{l=1}^{K}w_{il}^{t}(log(p_l(\boldsymbol{X_{i}}|\boldsymbol{\lambda_{il}},\Theta _l)))
\end{equation}
Expanding $p_{l}$ using equation~(3), we have:
\begin{equation}
\begin{split}
L_{2}(\Lambda_{1},...,\Lambda_{K},\Theta_{1},...,\Theta_{K})= \\
\sum_{i=1}^{n}\sum_{j=1}^{d}\sum_{l=1}^{K}-w_{il}^{t}e^{\lambda _{ijl}}+\sum_{i=1}^{n}\sum_{j=1}^{d}\sum_{l=1}^{K}w_{il}^{t}\lambda _{ijl}x_{ij}-\\
\sum_{i=1}^{n}\sum_{j=1}^{d}\sum_{l=1}^{K}w_{il}^{t}log(x_{ij}!)-\frac{d}{2}\sum_{i=1}^{n}\sum_{l=1}^{K}w_{il}^{t}log(2\pi)+\\\frac{1}{2}\sum_{i=1}^{n}\sum_{l=1}^{K}w_{il}^{t}log(det(\Sigma ^{-1}_{l}))-\\\frac{1}{2}\sum_{i=1}^{n}\sum_{l=1}^{K}w_{il}^{t}(\boldsymbol{\lambda _{il}}-\boldsymbol\mu _{l})^{T}\Sigma ^{-1}_{l}(\boldsymbol{\lambda _{il}}-\boldsymbol\mu _{l})
\end{split}
\end{equation}
where $\lambda_{ijl}$ is the $j^{th}$ entry in $\boldsymbol{\lambda_{il}}$. We solve for the parameters separately using the partial derivation method. 
Calculation of the partial derivative of $L_{2}$ with respect to $\lambda_{ijl}$ gives us:
\begin{equation}
\begin{split}
\frac{\partial }{\partial \lambda _{ijl}}L_{2}=0 \Rightarrow \\
-e^{\lambda _{ijl}} + x_{ij}-\inner{(\boldsymbol{\lambda _{il}}-\boldsymbol{\mu _{l}})}{\Sigma _{.jl}^{-1}}=0 
\end{split}
\end{equation}
where $\inner{\boldsymbol{a}}{\boldsymbol{b}}$ denotes the inner product of vectors $\boldsymbol{a}$ and $\boldsymbol{b}$, and $\Sigma _{.jl}^{-1}$ denotes the $j^{th}$ column of $\Sigma _{l}^{-1}$ . 
We use the Newton-Raphson method to estimate $\lambda_{ijl}^{t+1}$ from this equation, using values from iteration $t$ for $\boldsymbol{\Sigma^{-1}}$, $\boldsymbol{\mu}$, and ${\lambda_{ij'l}}$ (where $j \ne j'$).\\
Partial derivation with respect to $\boldsymbol{\mu_{l}}$ gives us:
\begin{equation}
\begin{split}
\frac{\partial }{\partial \boldsymbol\mu _{l}}L_{2}=0 \Rightarrow \\
\sum_{i=1}^{n}w_{il}^{t}(\boldsymbol{\lambda _{il}}-\boldsymbol{\mu _{l}})=0 \Rightarrow \\
\boldsymbol{\mu _{l}}=\frac{\sum_{i=1}^{n}w_{il}^{t}\boldsymbol{\lambda _{il}}}{\sum_{i=1}^{n}w_{il}^{t}}
\end{split}
\end{equation}
Thus, $\boldsymbol{\mu_l}^{t+1}$ can be estimated from $w_{il}^{t}$ and $\boldsymbol{\lambda_{il}^{t+1}}$.
\\\\
And finally, partial derivation with respect to $\Sigma^{-1}_{l}$ results in :
\begin{equation}
\begin{split}
\frac{\partial}{\partial \Sigma _{l}^{-1}}L_{2}=0 \Rightarrow \\
\frac{\partial }{\partial \Sigma _{l}^{-1}}  \{ \frac{1}{2}\sum_{i=1}^{N}w_{il}^{t}log(det(\Sigma _{l}^{-1}))-\\
\frac{1}{2}trace\left ( \Sigma _{l}^{-1}\sum_{i=1}^{N}w_{il}^{t}(\boldsymbol{\lambda _{il}}-\boldsymbol{\mu _{l}})(\boldsymbol{\lambda _{il}}-\boldsymbol{\mu _{l})}^{T} \right )  \} =0  
\end{split}
\end{equation}
We can solve this equation using matrix derivative rules to compute an estimate for the covariance matrix $\Sigma_{l}$ in iteration $(t+1)$ as:
\begin{equation}
\Sigma _{l}^{t+1}=\frac{\sum_{i=1}^{n}w_{il}^{t}(\boldsymbol{\lambda _{il}^{t+1}}-\boldsymbol{\mu _{l}^{t+1}})(\boldsymbol{\lambda _{il}^{t+1}}-\boldsymbol{\mu _{l}^{t+1}})^{T}}{\bar\sum_{i=1}^{n}w_{il}^{t}}
\end{equation}
\subsection{Inferring sparse networks using an $l_{1}$-penalty model}
We extended the MixMPLN framework by incorporating an $l_{1}$-norm penalty as follows:
\begin{equation}
p(\boldsymbol{X}|\pi_{1},...,\pi_{K},\Theta_{1},...,\Theta_{K})=\sum_{l=1}^{K}[\pi _{l}p_{l}(\boldsymbol{X}|\Theta _{l}) + \rho_{l} \left \| \Sigma_{l}^{-1}  \right \| _{1}]
\end{equation}
where $ \left \| \Sigma^{-1}_{l}\right \|_{1} $ is the $ l_{1}$-norm of the precision matrix of the $l^{th}$ component, and $\rho_{1},.., \rho_{K}$ are tuning parameters that can be selected independently. This framework allows for the inference of sparse networks associated with the $K$ components.
For a fixed tuning parameter of $\rho$ and a multivariate Gaussian distribution, the problem of selecting a precision matrix with the $l_{1}$-norm penalty can be stated as \cite{glasso}:
\begin{equation} 
\underset{\Sigma^{-1} }{argmax}\left \{ log(det(\Sigma^{-1}))-trace(S\Sigma^{-1})-\rho \left \| \Sigma^{-1} \right \| _{1} \right\}
\end{equation}
where $S$ denotes the empirical covariance matrix.\\
In our implementation of the extended MixMPLN framework, we calculated the empirical covariance matrix for each component using equation (20) in each iteration.  
We used the "glasso" and "huge" packages in R to solve the sparse precision matrix selection problem \cite{glasso,huge}. We applied three different strategies using each of the two packages. In MixMPLN+glasso(cross validation), we used cross validation to determine the value of $\rho$ (i.e. we picked that $\rho$ value which gave the best log-likelihood value among the subsamples). In MixMPLN+glasso(fixed tuning parameter), we used $\rho = 2d^2/(N\left \| \Sigma_{init}^{-1}  \right \| _{1})$, as proposed in \cite{biswas2015learning}. In this formulation, $\Sigma_{init}^{-1}$ is the estimated precision matrix after the initialization step of MixMPLN. In MixMPLN+glasso(iterative tuning parameter), we used the same formulation to initialize the $\rho$ value, but then updated it in each iteration based on the new estimated precision matrix in that iteration. In MixMPLN+huge(StARS), we used the stability approach to regularization selection (StARS) method. This method selects the coefficient which results in the most edge stability in the final graph \cite{SPIEC-EASI}. The tuning parameter selection method in MixMPLN+huge(fixed tuning parameter) and MixMPLN+huge(fixed tuning parameter) were the same as the corresponding implementations using glasso.  

\begin{figure*}[htp]
\includegraphics[width=1\linewidth]{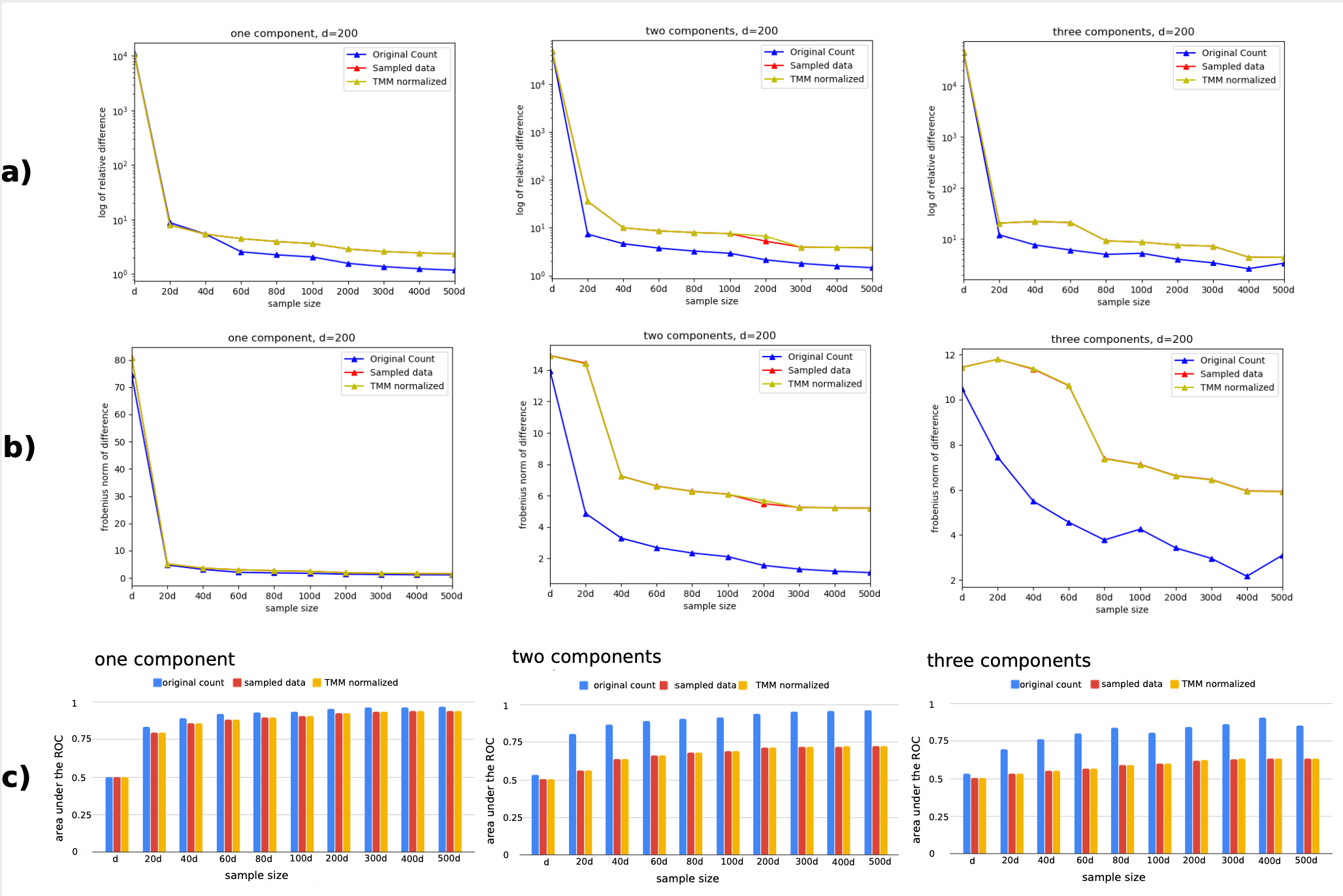} 
\caption{Performance on synthetic data with $d=200, sp=0.7, K=1,2,3$. a) Relative difference between the predicted and true precision matrices. b) Frobenius norm of the difference. c) Area under the ROC. }
\end{figure*} 

\begin{table}[htp]
\caption{Performance on synthetic data with $d=200, sp=0.9, K=1,2,3$.}
\begin{center}
  \def\arraystretch{.2}
  \begin{tabular*}{.75\textwidth} {p{6cm}L{1.6cm}C{1.6cm}C{1.6cm}}
  	\hline
   one component Frobenius norm  & n=2d & n=5d & n=10d\\ \hline
   MixMPLN &14.16 &13.99 &7.34\\
   MixMPLN+huge(StARS) &5.62 &16.02 &7.15\\
   MixMPLN+huge(fixed $\rho$) &4.86 &2.87 &2.08\\ 
   MixMPLN+huge(iterative $\rho$) &5.62 &3.01 &2.15\\
   MixMPLN+glasso(cross validation) &4.17 &2.96 &2.09\\
   MixMPLN+glasso(fixed $\rho$) &4.86 &2.87 &2.08\\ 
   MixMPLN+glasso(iterative $\rho$) &5.62 &3.01 &2.15\\ \hline
   
   one component Relative distance  & n=2d & n=5d & n=10d\\ \hline
   MixMPLN &27.36 &30.23 &11.75 \\
   MixMPLN+huge(StARS) &1.96 &30.31 &9.62  \\
   MixMPLN+huge(fixed $\rho$) &3.10 &1.04 &0.72  \\ 
   MixMPLN+huge(iterative $\rho$) &4.80 &1.49 &1.07 \\
   MixMPLN+glasso(cross validation) &1.52 &1.23 &0.77  \\
   MixMPLN+glasso(fixed $\rho$) &3.10 &1.04 &0.72  \\ 
   MixMPLN+glasso(iterative $\rho$) &4.80 &1.49 &1.07   \\ \hline
 
   two components Frobenius norm  & n=2d & n=5d & n=10d\\ \hline
   MixMPLN &73.67&19.76&8.79\\
   MixMPLN+huge(StARS) &5.60&5.42&15.29\\
   MixMPLN+huge(fixed $\rho$) &27.24&5.28&3.48\\ 
   MixMPLN+huge(iterative $\rho$) &25.94&4.33&2.88\\
   MixMPLN+glasso(cross validation) &5.04&3.73&2.84\\
   MixMPLN+glasso(fixed $\rho$) &28.83&5.28&3.48\\ 
   MixMPLN+glasso(iterative $\rho$) &27.63&4.33&2.88\\ \hline
 
   two components Relative distance  & n=2d & n=5d & n=10d\\ \hline
   MixMPLN &14443.03&56.52&14.42 \\
   MixMPLN+huge(StARS) &1.98&1.83&25.50 \\
   MixMPLN+huge(fixed $\rho$) &957.78&4.68&2.52  \\ 
   MixMPLN+huge(iterative $\rho$) &772.93&2.73&1.29 \\
   MixMPLN+glasso(cross validation) &1.82&1.30&1.04  \\
   MixMPLN+glasso(fixed $\rho$) &920.39&4.68&2.52  \\ 
   MixMPLN+glasso(iterative $\rho$) &842.48&2.73&1.29   \\ \hline
 
    three components Frobenius norm  & n=2d & n=5d & n=10d\\ \hline
    MixMPLN &20.07&14.66&13.61\\
    MixMPLN+huge(StARS) &6.93&5.90&5.46\\
    MixMPLN+huge(fixed $\rho$) &19.15&8.58&6.04\\ 
    MixMPLN+huge(iterative $\rho$) &18.56&6.11&4.97\\
    MixMPLN+glasso(cross validation) &6.48&6.25&5.88\\
    MixMPLN+glasso(fixed $\rho$) &19.67&11.02&8.03\\ 
    MixMPLN+glasso(iterative $\rho$) &19.44&6.11&4.98\\ \hline
 
    three components Relative distance  & n=2d & n=5d & n=10d\\ \hline
    MixMPLN &48131.48&4571.65&24.22\\
    MixMPLN+huge(StARS) &8959.39&2.18&1.87\\
    MixMPLN+huge(fixed $\rho$) &16096.08&12.68&6.70\\
    MixMPLN+huge(iterative $\rho$) &5891.79&5.63&3.04\\
    MixMPLN+glasso(cross validation) &2.60&2.46&3.55\\
    MixMPLN+glasso(fixed $\rho$) &11828.42&2029.51&10.91\\
    MixMPLN+glasso(iterative $\rho$) &5836.67&5.63&3.32\\ \hline

    \hline
  \end{tabular*}
 	\end{center}
\end{table}
\begin{figure}[htp]
\includegraphics[width=.7\linewidth]{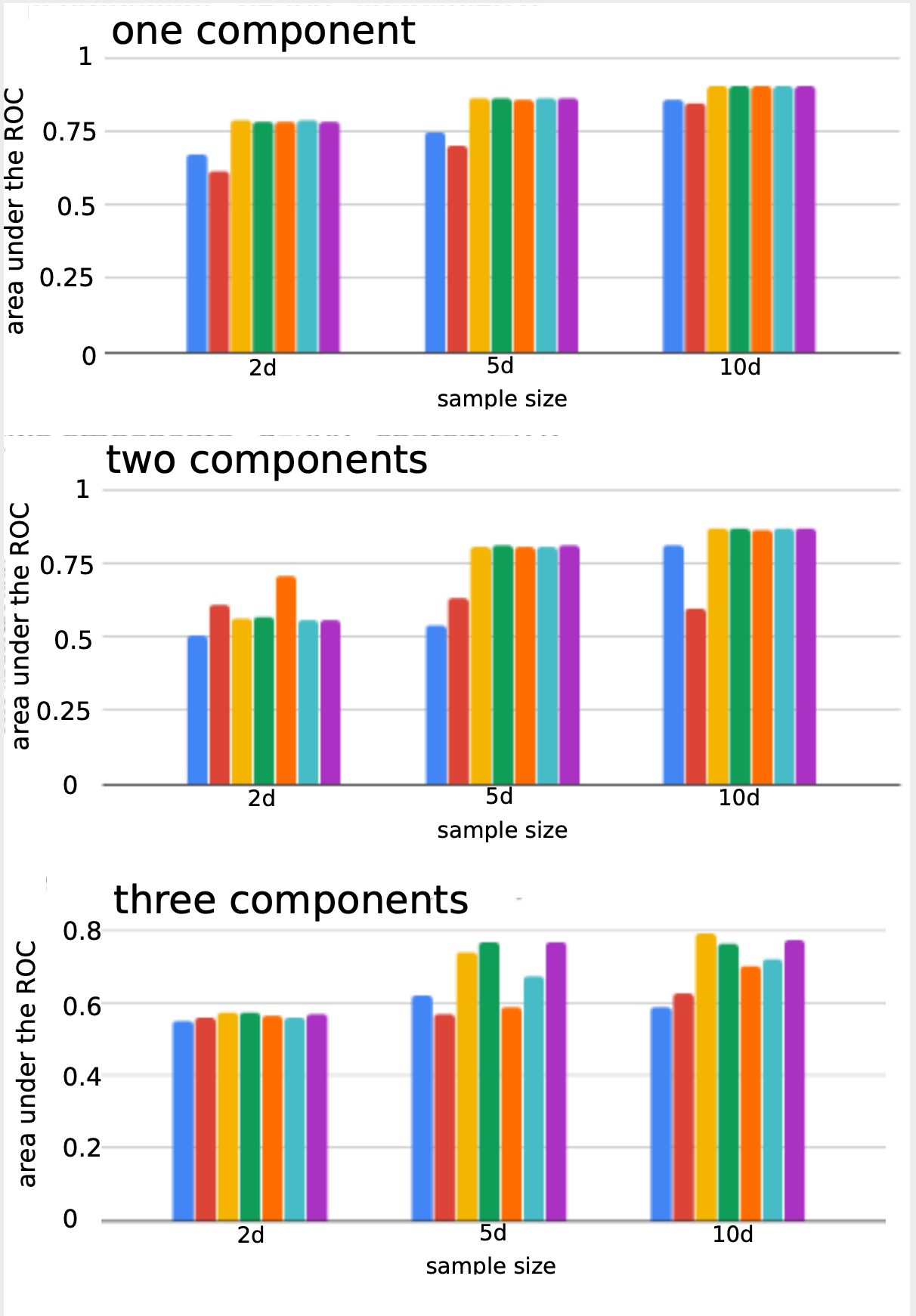}
\caption{AUC values for the synthetic datasets generated using $d=200, sp=0.9, K=1,2,3$.}
\end{figure} 
 
\begin{figure*}[htp]
    \centering
    \includegraphics[width=1\linewidth]{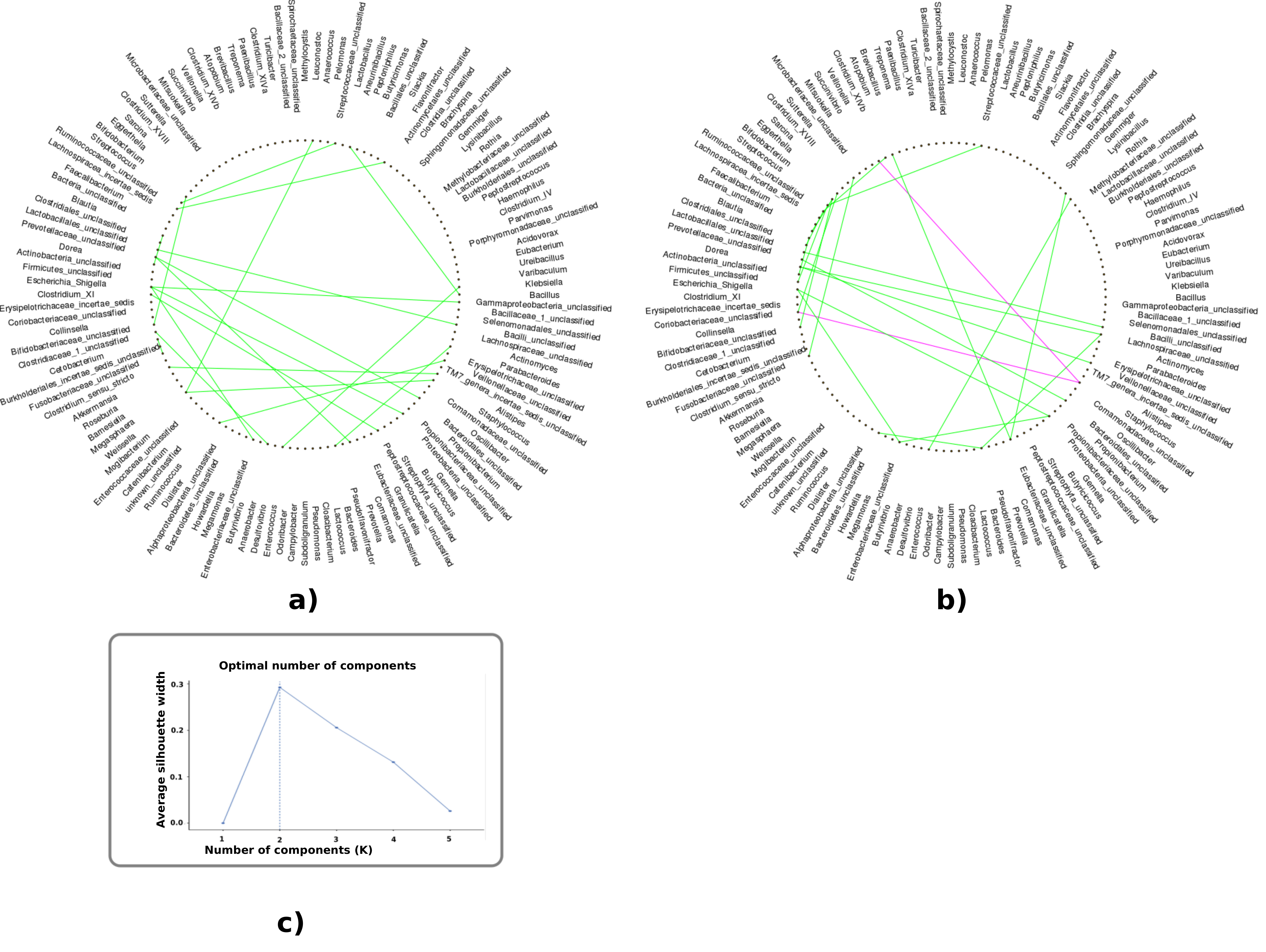}
    \caption{Application of MixMPLN+glasso with cross validation to a real dataset. Green and red edges represent positive and negative entries respectively in the estimated partial correlation matrices. a) Graph of Component 1 which contains 158 samples; b) graph of Component 2 which contains 37 samples. The threshold to select the edges is 0.3; c) Selection of the optimal number of the components.}
\end{figure*}
\begin{figure*}[htp]
    \centering
    \includegraphics[width=1\linewidth]{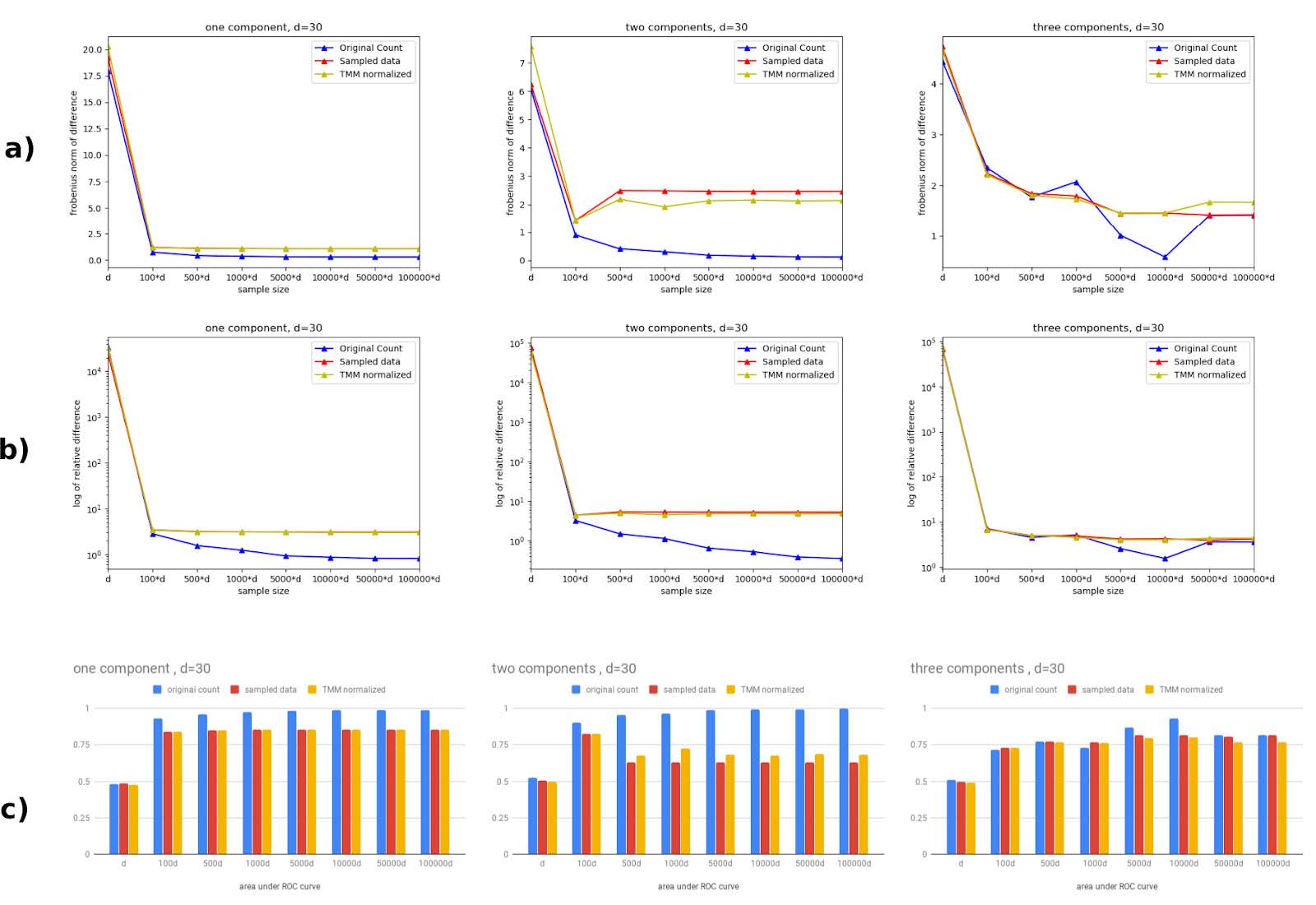}
    \caption{Result for synthetic data with d=30. a)relative difference between estimated precision matrix and the real one.b)Frobenius form of the difference. c)area under the ROC}
\end{figure*}
\begin{figure*}[htp]
    \centering
    \includegraphics[width=1\linewidth]{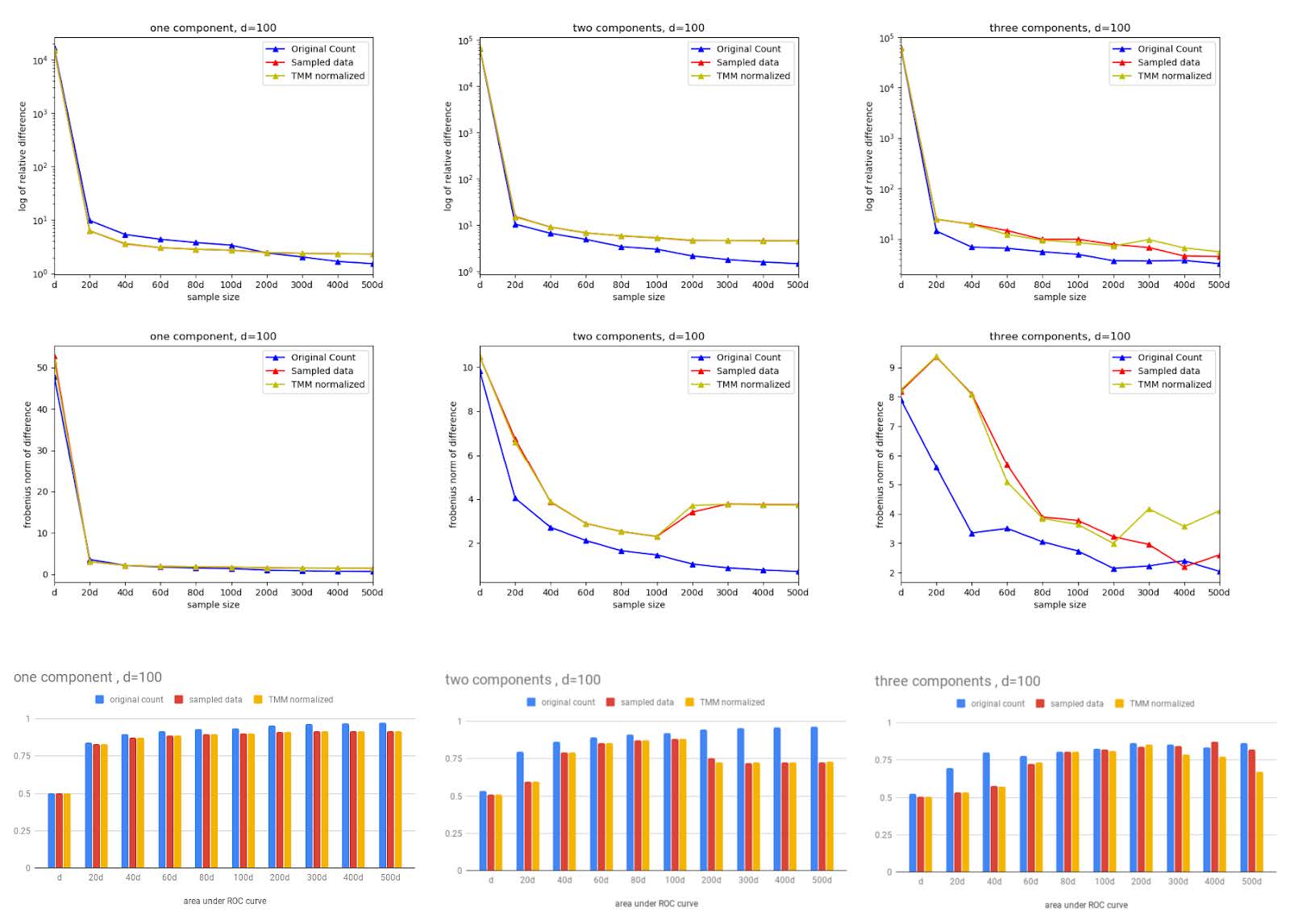}
    \caption{Result for synthetic data with d=100. a)relative difference between estimated precision matrix and the real one. b)Frobenius form of the difference. c)area under the ROC}
\end{figure*}
\begin{figure*}[htp]
    \centering
    \includegraphics[width=.845\linewidth]{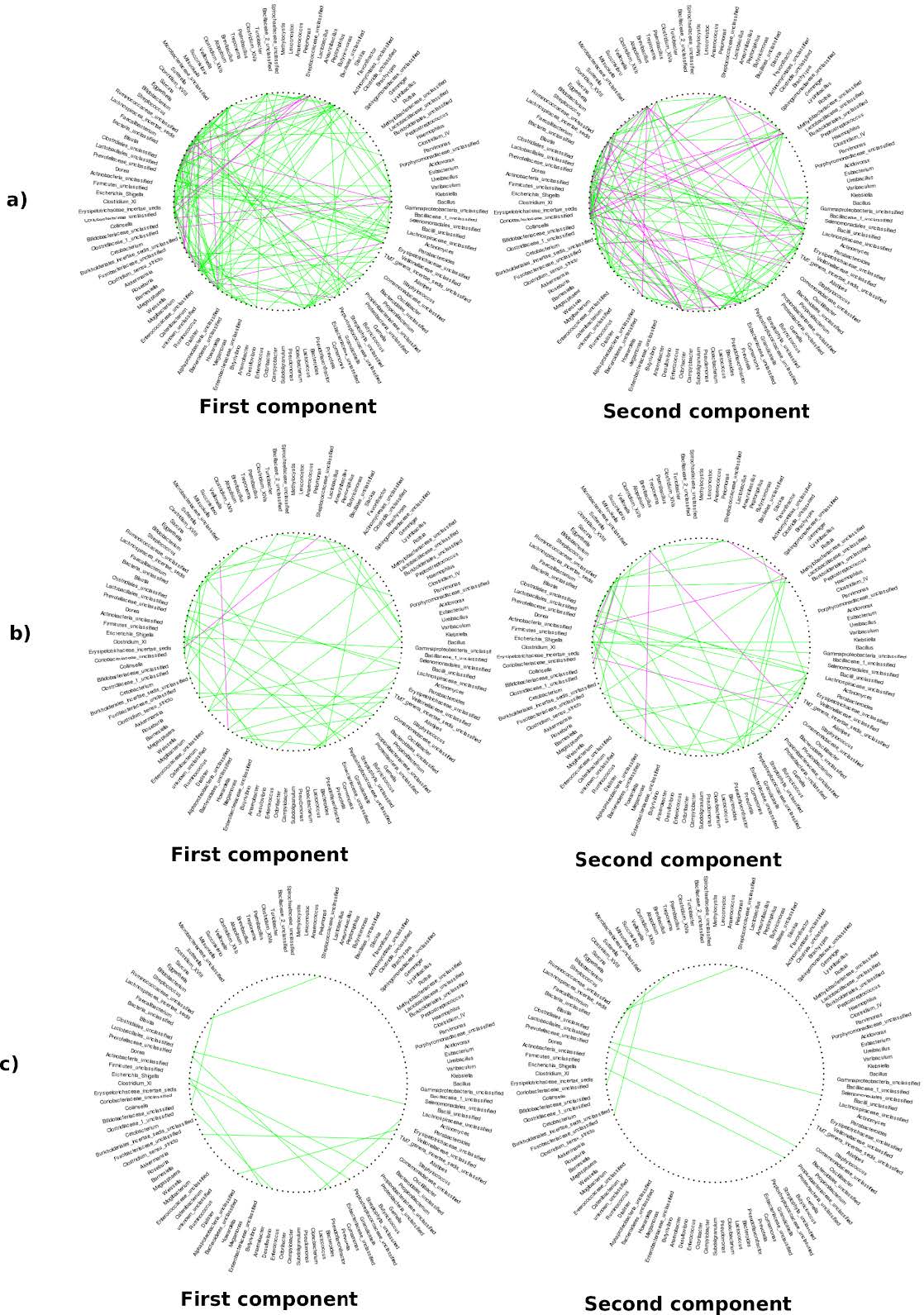}
    \caption{Applying MixMPLN+glasso with cross validation on real dataset. Green and red edges represent positive and negative entries respectively in the estimated partial correlation matrix. a) Threshold = 0.1 , b) Threshold = 0.2 , c)Threshold = 0.4}
\end{figure*}
\section{Performance evaluation and datasets}

Synthetic sample-taxa count matrices were generated in order to assess
the performance of the various MixMPLN algorithms. We evaluated
the convergence properties of the algorithms as a function of increasing
sample sizes. Since, in practice, sample-taxa count matrices generated
from biological samples are compositional in nature, we also evaluated
the effect of sampling from the true (or absolute) counts, and the subsequent
application of data normalization, on network recovery and
convergence. In addition, we also evaluated the accuracy in recovering sparse networks. Finally,
we applied our mixture model framework to analyze a real dataset.
\subsection{Datasets}

{\bf Synthetic data generation:}
Each sample-taxa count matrix $X$ was produced
by combining samples generated from $K$ component MPLN distributions,
where component $l$ generated $n_l$ samples ($d$-length count vectors), and
such that $\pi_{l}=\frac{n_l}{n}$ and $n=\Sigma_{l=1}^{K}n_{l}$. For
each component $l$, the $d \times d$ covariance matrix of its MPLN
distribution was derived from a randomly generated $d \times d$
positive definite precision matrix containing a fixed number of
zero entries (as given by the sparsity level $sp$ which denotes 
the fraction matrix entries that are zero). The mean
vector for each MPLN component was a random $d$-length
vector.  For a sample-taxa matrix $X$ generated this way, it is
assumed that each entry in the matrix is the true (or absolute) abundance of taxon $j$
in sample $i$. We refer to $X$ as the {\em orginal count}
matrix. To simulate compositional count data and sequencing depth
differences between the biological samples, we generated a {\em
sampled data} matrix $XS$ from $X$ by first normalizing each entry in a sample (by dividing by sample size)
and subsequently scaling that value by a sample-specific random number $Y \in
[5000,10000]$ (to model sequencing depth for the biological sample). Finally, to study the effect of data
normalization, we applied the trimmed mean of M-values (TMM)
normalization procedure (from the "edgeR" package \cite{TMM}) to $XS$ to
generate the {\em TMM normalized data} matrix $TS$.\\

\noindent{\bf Real data:} We also applied our mixture modeling framework to a sample-taxa count matrix produced by a
recent microbiome study \cite{yooseph2015stool} that explored connections between gut microbiome composition
and the risk of {\em Plasmodium falciparum} infection. In this study, stool samples from a cohort composed
of 195 Malian adults and children were collected and analyzed. The samples
were assayed by sequencing the 16S ribosomal RNA gene to determine the bacterial communities
they contained. This generated a sample-taxa count matrix with 195 samples and 221 bacterial genera which
we analyzed in this study.
\subsection{Benchmarking criteria}
Let $A_{d \times d}$ denote the true precision matrix and $B_{d \times d} $ an inferred (or predicted) precision matrix. For evaluating the performance of our algorithms on synthetic data, we used three different criteria to compare the inferred precision matrices with the original precision matrices that were used to generate the sample-taxa matrices. 
\begin{itemize}
\item {\em Relative difference} between two matrices $A$ and $B$ defined as $ \frac{1}{d^{2}}\sum_{i=1}^{d}\sum_{j=1}^{d}\frac{\left | a_{ij}-b_{ij} \right |}{\left | a_{ij}\right |.\left | b_{ij}\right |}$ . 
\item {\em Frobenius norm of the difference} between the partial correlations of matrices $A$ and $B$. Frobenius norm of a matrix $M$ is defined as $\left \| M \right \|_{F}=\sqrt{\sum_{i} \sum_{j}M_{ij}^{2} }$. For a precision matrix $C$, its partial correlation matrix $P$ is calculated as $P[i,j]=-C [i,j]/\sqrt{C[i,i]*C[j,j]}$.
\item {\em Area under the ROC (AUC)}: this measure was used to assess the recovery of the edges of the microbial network (i.e. identification of the zero entries in the precision matrix). The AUC was calculated by applying varying thresholds to the original and estimated precision matrices to define zero and non-zero entries. As any non-zero entry in the estimated precision matrix represents an edge in the  graph, the specificity and sensitivity of detecting an edge at different thresholds can be computed and used to plot the ROC. \vspace*{1pt}
\end{itemize}
For the above measures, smaller values for relative difference and frobenius norm indicate increased proximity to the ground truth. Values closer to 1 for the AUC plots indicate increased accuracy in reconstructing the network topology.
When $K>1$, we first matched the predicted precision matrix (of a component) to the nearest true precision matrix from the set of $K$ true precision matrices. We used the Frobenius norm measure for this. After pairing the true and predicted matrices, we report their mean value statistics.  

\section{Results}
We implemented the MixMPLN algorithms in R, and assessed their performance using the synthetic
datasets. For our assessments, we generated sample-taxa count matrices $X$ (and their corresponding $XS$ and $TS$ matrices) for the following four sets of parameters
\begin{itemize}
\item ($d=30,sp=0.5,n=[d,100000d],K=1,2,3$),
\item ($d=100,sp=0.7,n=[d,500d],K=1,2,3$),
\item ($d=200,sp=0.7,n=[d,500d],K=1,2,3$),
\item ($d=200,sp=0.9,d=2d,5d,10d, K=1,2,3$).
\end{itemize}

For each dataset, each component mixing coefficient was $1/K$. In addition, 5 replicates were generated for each parameter combination. In total, 465 datasets were generated and analyzed. The datasets with $sp=0.9$ were used to assess the performance of the MixMPLN algorithms in recovering sparse networks.

First we evaluated the ability of the MixMPLN algorithm to
recover the true precision matrices with increasing sample size $n$. For this, we used 
the original count data, the sampled data, and the TMM normalized
data. Figure 3.1 shows the benchmark results for the parameter combination of $d=200, sp=0.7$ and
$K=1,2,3$; Figure 3.4 and Figure 3.5 show the corresponding results for $d=30, sp=0.5, K=1,2,3$, and for $d=100, sp=0.7, K=1,2,3$. The three benchmark criteria (relative difference,
Frobenius norm, and AUC) show that the entries in predicted precision matrices
approach their true values as the sample size $n$ increases. 
A strong convergence trend is seen using the original count data
(blue curves/barchart), with the AUC values approaching 0.97, 0.96,
and 0.9 for $K=1,2,3$ respectively. The drop in performance with increasing $K$ is
not unexpected given the smaller fraction of data available per
component to infer the component parameters. The figure also shows that
the accuracy of recovery of the true precision matrices is not as high when using
the sampled data (red curves/barchart) and the TMM normalized data
(orange curves/barchart). In addition, from our analysis, it is not immediately evident 
that applying a TMM type data normalization is advantageous
for the purpose of inferring the underlying covariance structure of the network.

We also evaluated the performance of MixMPLN and its  $l_{1}$-norm penalty variants
in recovering sparse networks. Table 3.1 and Figure 3.2 show the
performance of these methods for $d=200$, $K=1,2,3$, and sparsity
level $sp=0.9$. Sample sizes of $n=2d, 5d, 10d$ were used in these
evaluations. As can be seen, the performance for the methods generally
improve with increasing $n$, and the  $l_{1}$-norm penalty variants perform
better than the unpenalized MixMPLN model on these data. The performances of the  $l_{1}$-norm penalty
variants are generally quite comparable to each other (with
MixMPLN+huge(StARS) having a slightly lower performance compared to the
others). 

Since MixMPLN+glass(cross validation) performs better than the other approaches for $n=2d$, we decided to apply this method to analyze the real dataset. We used the silhouette method from the "factoextra" R package \cite{factoextra} to compute the optimal number of components from this sample-taxa matrix. Figure 3.3 shows the results of our analysis. We find that there is strong evidence for two underlying (and different) microbial networks ($K=2$ components) associated with this sample-taxa data. We assigned component membership to the samples based on their final weights $w_{il}$ (Equation 3.10). This resulted in component 1 containing 158 samples and component 2 containing 37 samples. The average age of the individuals in component 1 was 9 years while that of individuals in component 2 was 0.7 years. Our reconstructed networks are consistent with the observation that infants have a different gut microbiome composition compared to older children and adults \cite{yooseph2015stool}. The constructed networks include edges involving bacterial groups like {\em Bifidobacterium, Staphylococcus, Streptococcus}, and {\em Escherichia-Shigella}, that are known to be key players in early gut microbiome development. Our method identifies both positive and negative interactions between pairs of taxa (red and green edges) for the chosen threshold of 0.3; Figure 3.6 shows the graph structures for other selected threshold values. The biological significance of these interactions need to be investigated further. 

\section{Summary}
In this chapter, we presented a mixture model framework and network inference algorithms to analyze sample-taxa matrices that are associated with $K$ microbial networks. Next two chapters will include two other mixture model frameworks and network inference algorithms to analyze sample-taxa matrices.  

\chapter{MCMC FRAMEWORK TO MODEL THE DATA WITH MIXTURE OF MPLN}

\section{Proposed method}

The difference between MixMCMC and MixMPLN is in the stage of the estimation of the latent variables. Both MixMPLN and MixMCMC model the sample-taxa matrix by a mixture of MPLN distributions and use the MM strategy to simplify the log likelihood function. Although MixMPLN uses the gradient ascent to estimate all the parameters($\Theta$ and $\lambda$ for each component), MixMCMC use the gradient ascent to estimate $\Theta$ along with Markov Chain Monte Carlo (MCMC) sampling to estimate the latent variables ($\lambda$). In MixMCMC instead of solving equation~(12) to update $\boldsymbol{\Lambda}$, a sampling-based strategy is used. \\
MCMC is a general framework that allows sampling from a target distribution, and can be used to estimate parameters when the target distribution is the posterior distribution. Second MC refers to Monte Carlo which is to find the approximate by averaging. The first MC is Mrakov Chain which is a mechanism to generate samples. A sequence $X_{0},X_{1},...,X_{n}$ of random variables is a Markov Chain if, for any $t$, the conditional distribution of $X_{t}$, given $x_{t-1},x_{t-2},...,x_{0}$ is the same as the distribution of $X_{t}$ given $x_{t-1}$. In MCMC, we start navigating through the space and reject or accept the new sample until the convergence. \\
In our problem, the posterior distribution is $P(\boldsymbol{\lambda_{l}}|X)$, the probability of the latent variable($\boldsymbol{\lambda_{l}}$) given the observed sample-taxa count matrix($X$):
\begin{equation}
    P(\boldsymbol{\lambda_{l}}|X)=\frac{P(X|\boldsymbol{\lambda_{l}})P(\boldsymbol{\lambda_{l}})}{P(X)}
\end{equation}
 We used the {\em Rstan} package \cite{carpenter2017stan} to implement the MixMCMC akgorithm. In iteration $t+1$, the prior distribution of $\boldsymbol{\lambda_{l}}$ to feed the MCMC is a Multivariate Gaussian distribution with mean $\boldsymbol{\mu_{l}^{t}}$ and covariance matrix of $\Sigma_{l}^{t}$ (where $l$ represents the component number), and the likelihood function $P(X|\boldsymbol{\lambda_{l}})$ for sampling $\boldsymbol{\lambda_{l}}$ follows an MPLN distribution.

For each MCMC execution, we ran three chains in parallel \cite{Annis2016}. The $\hat{R}$ (between-chain variance relative to that of within-chain variance) \cite{Gelman1992} and $n_{eff}$ (the effective sample size) \cite{Gelman2013} values were used to check convergence of these chains. While running the MCMC for each chain, we started with 1000 iterations and set half of them as warm-up iterations. After each running, convergence was assumed if $\hat{R}$ was less than 1.1 and $n_{eff}$ was higher than 100. In the case that the MCMC had not converged, we increased the number of iterations by 100 and ran the chains again. We repeated this process until convergence.\\
Aside from the mentioned difference (using the MCMC to estimate the latent variables of lambda), the other steps including applying CV to find the tuning parameter for the sparsity constraint in MixMCMC and MixMPLN are the same.
 \begin{figure*}
    \centering
    \includegraphics[width=\linewidth]{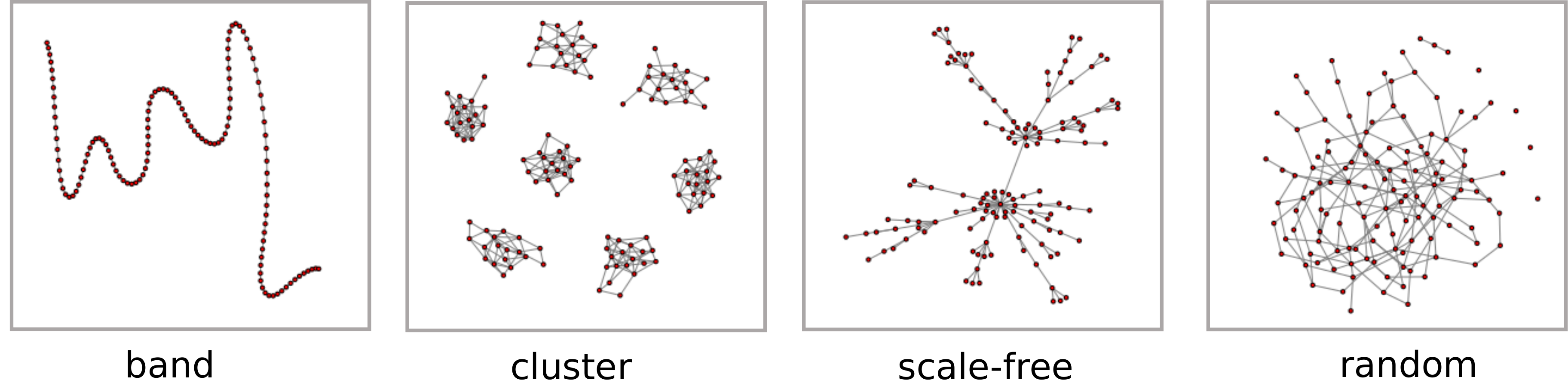}
    \caption{The different classes of graphs that our methods are evaluated on.}
\end{figure*}  

\section{Synthetic dataset generation}
Ground truth datasets from real microbial communities containing information on microbial interactions, and that could be used to assess the performance of network inference methods, are currently lacking in this field. Thus, we generated realistic synthetic datasets using the Normal To Anything (NorTA) framework \cite{cario1997modeling} \cite{kurtz2015sparse}. NorTA is a method which allows sample generation with arbitrary marginal distributions and correlation matrices. NorTA applies the inverse CDF of the target distribution to the CDF of the normal distribution with the desired correlation matrix. We used NorTA to generate datasets which follow zero-inflated Negative Binomial distribution with the same statistical features of the gut microbiome data from the Human Microbiome Project (HMP)  \cite{jumpstart2012evaluation}\cite{kurtz2015sparse}\cite{decode}.

We used the {\em huge} package to generate three different graph types (band, cluster and scale-free) \cite{barabasi1999emergence}; we also generated random graphs using our previous approach \cite{ISMB}. Each generated graph (Figure 4.1) gives us a correlation matrix as an input for the NorTA method. For producing datasets for multiple components (i.e. when $K>1$), the graph for each component is derived from the graph for the first component by relabeling the nodes using a random permutation. The correlation matrix for each component is fed to the NorTA method and used to generate the samples corresponding to that component. Subsequently, the samples for the different components are combined to generate a mixture dataset. The final step in generating the synthetic data is to mimic the sampling process by first normalizing each entry in a sample (by dividing by sample size) and subsequently scaling that value by a sample-specific random number
between 5000 and 10000 (to model sequencing depth for the biological sample).

In our simulations, all generated graphs have $90\%$ sparsity (i.e. only $10\%$ of the possible edges are present). We used an HMP 16S rRNA dataset \cite{jumpstart2012evaluation} to find the statistical features of real data, and thus $d$, the number of taxa, in our simulation is the same as the number of taxa in that dataset ($d = 127$). Synthetic data corresponding to mixtures with $K=1, 2,$ and $3$ components were generated, and used to assess the performance of our methods. For each evaluation, we generated three different sample sizes ($d, 2d,$ and  $3d$) for each component, to reflect sample sizes often available in practice. For each evaluation, five replicate datasets were generated, and the final performance numbers reported are the average values of the five replicates.


\begin{table}[htp]
\caption{AUC performance on synthetic data with "band" graph type with $d=127$. "$K$" indicates the number of components to generate the synthetic data. "$n$" is the number of samples in each component. Running time is indicated next to the AUC while s, h and d are abbreviations for seconds, hours and days respectively.}
\begin{center}
   
  \def\arraystretch{.5}
\begin{tabular*}{\textwidth} { |p{3cm}L{3cm}L{2.6cm}L{2.6cm}L{3.2cm}|}  	\hline
   &      &   &  &  \\
  \textbf{type} & \textbf{method}  & \textbf{n=d} & \textbf{n=2d} & \textbf{n=3d}\\   
   &      &   &  &  \\
   \hline
   \cellcolor{gray!20}&      &   &  &  \\
   \cellcolor{gray!20}&MixMPLN      &0.95 (6m) &0.97 (13m) &0.98 (6m)\\
   \textbf{K=1} \cellcolor{gray!20}&      &   &  &  \\
   \cellcolor{gray!20}&MixMCMC    &0.96 (20h) &0.97 (1.2d) &0.98 (1.6h)\\
   \cellcolor{gray!20}&      &   &  &  \\
   \hline
   \cellcolor{gray!20}&      &   &  &  \\
   \cellcolor{gray!20}&MixMPLN      &0.77 (12m) &0.88 (1.7h) &0.94 (1.3h)\\
   \textbf{K=2}\cellcolor{gray!20}&      &   &  &  \\
    \cellcolor{gray!20}&MixMCMC    &0.76 (2d) &0.85 (2.4d) &0.92 (3d)\\
   \cellcolor{gray!20}&      &   &  &  \\
   \hline
    \cellcolor{gray!20}&      &   &  &  \\
   \cellcolor{gray!20}&MixMPLN      &0.66 (1.4h) &0.75 (1.6h) &\textbf{0.98} (3.5h)\\
   \textbf{K=3}\cellcolor{gray!20}&      &   &  &  \\
    \cellcolor{gray!20}&MixMCMC    &0.64 (3.1d) &0.72 (4d) &0.97 (6.2d)\\
   \cellcolor{gray!20}&      &   &  &  \\

    \hline
    \end{tabular*}
 	\end{center}
\end{table}  
\begin{table}[htp]
\caption{AUC performance on synthetic data with "cluster" graph type with $d=127$. "$K$" indicates the number of components to generate the synthetic data. "$n$" is the number of samples in each component. Running time is indicated next to the AUC while s, h and d are abbreviations for seconds, hours and days respectively.}
\begin{center}
   
  \def\arraystretch{.5}
  \begin{tabular*}{\textwidth} { |p{3cm}L{3cm}L{2.6cm}L{2.6cm}L{3.2cm}|}
  \hline
  	 &      &   &  &  \\
  \textbf{type} & \textbf{method}  & \textbf{n=d} & \textbf{n=2d} & \textbf{n=3d}\\ 
   &      &   &  &  \\
   \hline
   \cellcolor{gray!20}&      &   &  &  \\
   \cellcolor{gray!20}&MixMPLN      &0.87 (14m) &0.92 (10.5m) &0.93( 13m)\\
   \textbf{K=1}\cellcolor{gray!20}&      &   &  &  \\
    \cellcolor{gray!20}&MixMCMC    &0.88 (1d) &0.89 (1.3d) &0.91 (2d)\\
   \cellcolor{gray!20}&      &   &  &  \\
    \hline
    \cellcolor{gray!20}&      &   &  &  \\
   \cellcolor{gray!20}&MixMPLN      &0.70 (48m) &0.76 (42m) &0.80 (3h)\\
   \textbf{K=2}\cellcolor{gray!20}&      &   &  &  \\
    \cellcolor{gray!20}&MixMCMC    &0.71 (2.3d) &0.73 (3d) &0.78 (3.3d)\\
   \cellcolor{gray!20}&      &   &  &  \\
    \hline
    \cellcolor{gray!20}&      &   &  &  \\
   \cellcolor{gray!20}&MixMPLN      &0.64 (1.1h) &0.70 (45m) &0.75 (2.3h)\\
   \textbf{K=3}\cellcolor{gray!20}&      &   &  &  \\
    \cellcolor{gray!20}&MixMCMC    &0.64 (3d) &0.70 (4.2d) &0.75 (6d)\\
   \cellcolor{gray!20}&      &   &  &  \\
   
    \hline
    \end{tabular*}
 	\end{center}
\end{table}  
\begin{table}[htp]
\caption{AUC performance on synthetic data with "random" graph type with $d=127$. "$K$" indicates the number of components to generate the synthetic data. "$n$" is the number of samples in each component. Running time is indicated next to the AUC while s, h and d are abbreviations for seconds, hours and days respectively.}
\begin{center}
   
  \def\arraystretch{.5}
\begin{tabular*}{\textwidth} { |p{3cm}L{3cm}L{2.6cm}L{2.6cm}L{3.2cm}|}  \hline
  	&      &   &  &  \\
  \textbf{type} & \textbf{method}  & \textbf{n=d} & \textbf{n=2d} & \textbf{n=3d}\\ 
  &      &   &  &  \\
   \hline
   \cellcolor{gray!20}&      &   &  &  \\
   \cellcolor{gray!20}&MixMPLN      &0.57 (6m) &0.62 (10m) &0.65 (6m)\\
   \textbf{K=1}\cellcolor{gray!20}&      &   &  &  \\
    \cellcolor{gray!20}&MixMCMC    &0.57 (18h) &0.58 (1.4d) &0.64 (2d)\\
   \cellcolor{gray!20}&      &   &  &  \\

    \hline
    \cellcolor{gray!20}&      &   &  &  \\
   \cellcolor{gray!20}&MixMPLN      &0.53 (1.2h) &0.55 (45m) &0.55 (2.1h)\\
   \textbf{K=2}\cellcolor{gray!20}&      &   &  &  \\
    \cellcolor{gray!20}&MixMCMC    &0.54 (2.3d) &0.53 (3d) &0.54 (3.8d)\\
   \cellcolor{gray!20}&      &   &  &  \\

    \hline
    \cellcolor{gray!20}&      &   &  &  \\
   \cellcolor{gray!20}&MixMPLN      &0.51 (1h) &0.53 (1h) &0.53 (3h)\\
   \textbf{K=3}\cellcolor{gray!20}&      &   &  &  \\
    \cellcolor{gray!20}&MixMCMC    &0.50 (3.5d) &0.52 (4.1d) &0.53 (5d)\\
   \cellcolor{gray!20}&      &   &  &  \\

    \hline
    \end{tabular*}
 	\end{center}
\end{table}  
\begin{table}[htp]
\caption{AUC performance on synthetic data with "scale-free" graph type with $d=127$. "$K$" indicates the number of components to generate the synthetic data. "$n$" is the number of samples in each component. Running time is indicated next to the AUC while s, h and d are abbreviations for seconds, hours and days respectively.}
\begin{center}
   
  \def\arraystretch{.5}
\begin{tabular*}{\textwidth} { |p{3cm}L{3cm}L{2.6cm}L{2.6cm}L{3.2cm}|}  \hline
  	&      &   &  &  \\
  \textbf{type} & \textbf{method}  & \textbf{n=d} & \textbf{n=2d} & \textbf{n=3d}\\ 
  &      &   &  &  \\
   \hline
   \cellcolor{gray!20}&      &   &  &  \\
   \cellcolor{gray!20}&MixMPLN      &0.64 (6m) &0.75 (6m) &0.83 (10m)\\
   \textbf{K=1}\cellcolor{gray!20}&      &   &  &  \\
    \cellcolor{gray!20}&MixMCMC    &0.64 (21h) &0.73 (1d) &0.80 (1.3d)\\
   \cellcolor{gray!20}&      &   &  &  \\
 
    \hline
    \cellcolor{gray!20}&      &   &  &  \\
   \cellcolor{gray!20}&MixMPLN      &0.58 (28m) &0.64 (42m) &0.63 (2h)\\
   \textbf{K=2}\cellcolor{gray!20}&      &   &  &  \\
    \cellcolor{gray!20}&MixMCMC    &0.60 (2d) &0.63 (2.4d) &0.61 (3.2d)\\
   \cellcolor{gray!20}&      &   &  &  \\
 
    \hline
    \cellcolor{gray!20}&      &   &  &  \\
   \cellcolor{gray!20}&MixMPLN      &0.53 (1h) &0.57 (2.2h) &0.57 (4h)\\
   \textbf{K=3}\cellcolor{gray!20}&      &   &  &  \\
    \cellcolor{gray!20}&MixMCMC    &0.53 (3.5d) &0.52 (4.5d) &0.54 (6d)\\
   \cellcolor{gray!20}&      &   &  &  \\
 
    \hline
    \end{tabular*}
 	\end{center}
\end{table}  
 
\section{Results}
\subsection{Results on synthetic data}

We used the synthetic data generated from the different graph types for various values of $K, d,$ and $n$ to evaluate the methods with sparsity constraints . For each dataset, every method was run three times with different random starting points, and the solution with the largest log-likelihood value was selected for that method. All methods were assessed with respect to their ability to recover the original graph topology, and thus, the area under the ROC curve (AUC) was used as the performance evaluation criterion. In addition, we also evaluated the running times of the methods. Experiments were carried out using an Intel(R) Xeon(R) E5-2640 v4@2.40GHZ machine.

The performance results on band, cluster, random and scale-free graphs are given in Tables 4.1, 4.2, 4.3 and 4.4 respectively. In the tables, the best AUC value (with at-least $2\%$ difference) in each experiment is shown in bold. Several patterns are noteworthy: (a) For each graph type and each $K$, as expected, the AUC value for a method increases with increasing $n$; this is a reflection of improved inference capability of each method with increasing sizes of the input data. (b) The ability of a method to recover the original graph is a function of the graph type. In general, the methods tend to be able to recover the topologies for band graphs and cluster graphs more accurately in comparison to topologies for scale-free graphs and random graphs (this is reflected in the comparatively higher AUC values for band and cluster graphs for each method), and MixMPLN and MixMCMC are comparable in all the cases with different graph types. In terms of running time, MixMCMC is significantly slower compared to the direct gradient ascent based approaches.

\section{Summary}
We introduced MixMCMC to infer multiple networks from microbial count data. It is based on the MM principle applied in a mixture model framework and build off of the MixMPLN framework that we introduced previously\cite{decode2}. We used synthetic datasets to evaluate these approaches with respect to recovering the original graph topologies, for different types of graphs. While the AUC values increase with increasing sample size of the input data, we observe that this is a function of the graph type. The methods tend to be able to recover the topologies for band graphs and cluster graphs more accurately in comparison to topologies for scale-free graphs and random graphs. However, MixMCMC is significantly slower.  

\chapter{DEALING WITH COMPOSITIONALITY FEATURE OF MICROBIAL DATA  }
In this chapter, modeling the data by Mixture of Graphical Gaussian (MixGGM) is introduced to deal with the compositionality feature in microbial data. 
\section{MixGGM algorithm}
 \noindent{\bf Data transformation using Centered Log Ratio transformation (CLR):} The first step in MixGGM involves applying the Centered Log Ratio (CLR) transformation to the sample-taxa matrix. CLR transformation is applied to each sample $\boldsymbol{X}$ to generate transformed data $\boldsymbol{Z}$ as follows: 
\begin{equation}
    \boldsymbol{Z}=clr(\boldsymbol{X})=[log(\frac{x_{1}}{g(\boldsymbol{X})}),...,log(\frac{x_{d}}{g(\boldsymbol{X})})]
\end{equation}
where $g(\boldsymbol{X})=\sqrt[d]{\prod_{i=1}^{d}x_{i}}$ is the geometric mean.

Let matrix $W$ denote the {\em absolute} abundances of the taxa in the samples. Suppose that $\Omega  =Cov(log(W))$ is the covariance of the {\em log} of the absolute count data that we are interested in estimating, and $ \Gamma =Cov(clr(X))$ is the covariance of $X$ after applying the CLR transformation to it. Then, these covariance matrices are related as follows  \cite{kurtz2015sparse}: 
\begin{equation}
\Gamma =G\Omega G    
\end{equation}
where $G=I_{d}-\frac{1}{d}J$, $I_{d}$ is a $d \times d$ identity matrix and $J$ is a $d \times d$ matrix with all 1's. When $d$ is large, $G$ is close to the identity matrix, so an approximation of $\Omega$ by $\Gamma$ is reasonable. For the MixGGM approach, we assume that the transformed data $Z$ follows a Multivariate Gaussian distribution \cite{aitchison1982statistical}.

\noindent{\bf Mixture of $K$ Gaussian distributions:} MixGGM models the CLR transformed data $Z$ by a mixture of Multivariate Gaussian distributions, which is defined by following equation:
\begin{equation}
p(Z|\pi_{1},...,\pi_{K},\Phi_{1},...,\Phi_{K})= 
\prod_{i=1}^{n}\sum_{l=1}^{K}\pi _{l}p_{l}(\boldsymbol{Z_{i}} |\Phi _{l})
\end{equation}
Where $p_{l}(\boldsymbol{Z_{i}} |\Phi _{l})$ follows the normal distribution:
\begin{equation}
(z_{1},...,z_{d})^{T} \sim \mathbb{N}_{d}(\boldsymbol{\mu} ,\Sigma)
\end{equation}
Therefore, the general log-likelihood function is: 
\begin{equation}
\begin{split}
L(\pi_{1},...,\pi_{K} ,\Phi_{1},...,\Phi_{K}|Z) = \\ log(\prod_{i=1}^{n}\sum_{l=1}^{K}\pi _{l}p_{l}(\boldsymbol{Z_{i}} |\Phi _{l}))
\end{split}
\end{equation}
where $\Phi_{l}$ includes $\boldsymbol{\mu}$ and $\Sigma$ for the $l^{th}$ component. We also use the MM framework to maximize the log-likelihood function $L$ and estimate the MixGGM parameters.  

\noindent{\bf Minorizer function :} The first step in applying the MM framework is to find a minorizer. We use equation~(3.9) to find the minorizer for function in equation~(5.5):
\begin{equation}
log(\prod_{i=1}^{n}\sum_{l=1}^{K}\pi _{l}p_{l}(\boldsymbol{Z_{i}} | \Phi _{l}))\geqslant\sum_{i=1}^{n}\sum_{l=1}^{K}w_{il}^{t}log(\frac{\pi_{l}}{w_{il}^{t}}p_{l}(\boldsymbol{Z_{i}}|  \Phi _{l}))
\end{equation}
where, weight $w_{il}^{t}$ is defined as follows:
\begin{equation}
w_{il}^t=\frac{\pi _{l}^{t} p_l(\boldsymbol{Z_{i}}|\Phi _l^t)}{\sum_{l=1}^{K}\pi _{l}^{t} p_l(\boldsymbol{Z_{i}}|\Phi _l^t)}
\end{equation}
 
\noindent{\bf Parameter estimation:} The process for updating the mixing coefficients, the mean vector and the covariance matrix for each component in iteration $t+1$ is similar to that employed for the MixMPLN algorithm \cite{ISMB}: 
\vspace{-.2cm}
\begin{equation}
\pi _{l}^{t+1}=\frac{1}{n}\sum_{i=1}^{n}w_{il}^{t}
\end{equation}
\vspace{-.2cm}
\begin{equation}
\boldsymbol{\mu _{l}^{t+1}}=\frac{\sum_{i=1}^{n}w_{il}^{t}\boldsymbol{Z_{i} }}{\sum_{i=1}^{n}w_{il}^{t}}
\end{equation}
\vspace{-.2cm}
\begin{equation}
\Sigma _{l}^{t+1}=\frac{\sum_{i=1}^{n}w_{il}^{t}(\boldsymbol{Z_{i}}-\boldsymbol{\mu _{l}^{t+1}})(\boldsymbol{Z_{i}}-\boldsymbol{\mu _{l}^{t+1}})^{T}}{\sum_{i=1}^{n}w_{il}^{t}}
\end{equation}
\noindent{\bf Applying sparsity constraint:} The solution to the problem of inferring a sparse precision matrix for a Multivariate Gaussian model when we have the empirical precision matrix ($S$) and tuning parameter ($\rho_{l}$) for the $l_{1}$-norm has been proposed previously (the graphical lasso) \cite{friedman2008sparse}. The graphical lasso formulation is as follows:
\vspace {-.3 mm}
\begin{equation} 
\underset{\Sigma^{-1} }{argmax}\left \{ log(det(\Sigma^{-1}))-trace(S\Sigma^{-1})-\rho \left \| \Sigma^{-1} \right \| _{1} \right\}
\end{equation}
In the MixGGM approach, this constraint is applied separately for all $K$ components. We used the {\em huge} package \cite{zhao2012huge} in our implementation to solve the graphical lasso. In our previous work \cite{ISMB}, several strategies were evaluated to select the tuning parameter. Based on those results, we use Cross Validation (CV) strategy for parameter tuning. The CV is based on dividing the data into random subgroups, and selecting the parameter which results in the best likelihood value in these comparisons. 

\begin{table}[htp]
\caption{AUC performance on synthetic data with "band" graph type with $d=127$. "$K$" indicates the number of components to generate the synthetic data. "$n$" is the number of samples in each component. Running time is indicated next to the AUC while s, h and d are abbreviations for seconds, hours and days respectively.}
\begin{center}
   
  \def\arraystretch{.5}
\begin{tabular*}{\textwidth} { |p{3cm}L{3cm}L{2.6cm}L{2.6cm}L{3.2cm}|}  	\hline
   &      &   &  &  \\
  \textbf{type} & \textbf{method}  & \textbf{n=d} & \textbf{n=2d} & \textbf{n=3d}\\   
   &      &   &  &  \\
   \hline
   \cellcolor{gray!20}&      &   &  &  \\
   \cellcolor{gray!20}&MixMPLN      &0.95 (6m) &0.97 (13m) &0.98 (6m)\\
\textbf{K=1}\cellcolor{gray!20}&      &   &  &  \\ 
   \cellcolor{gray!20}&MixGGM   &0.95 (6m) &0.97 (4.3m) &0.97 (4m)\\
   \cellcolor{gray!20}&      &   &  &  \\
    \hline
   \cellcolor{gray!20}&      &   &  &  \\
   \cellcolor{gray!20}&MixMPLN      &0.77 (12m) &0.88 (1.7h) &0.94 (1.3h)\\
   \textbf{K=2}\cellcolor{gray!20}&      &   &  &  \\
   \cellcolor{gray!20}&MixGGM  &\textbf{0.79} (48m) &\textbf{0.91} (38m) &\textbf{0.98} (22m)\\
   \cellcolor{gray!20}&      &   &  &  \\
    \hline
    \cellcolor{gray!20}&      &   &  &  \\
   \cellcolor{gray!20}&MixMPLN      &0.66 (1.4h) &0.75 (1.6h) &\textbf{0.98} (3.5h)\\
   \textbf{K=3}\cellcolor{gray!20}&      &   &  &  \\
   \cellcolor{gray!20}&MixGGM   &\textbf{0.73} (2h) &\textbf{0.80} (1.2h) &0.96 (1h)\\
   \cellcolor{gray!20}&      &   &  &  \\
   
    \hline
    \end{tabular*}
 	\end{center}
\end{table}  
\begin{table}[htp]
\caption{AUC performance on synthetic data with "cluster" graph type with $d=127$. "$K$" indicates the number of components to generate the synthetic data. "$n$" is the number of samples in each component. Running time is indicated next to the AUC while s, h and d are abbreviations for seconds, hours and days respectively.}
\begin{center}
   
  \def\arraystretch{.5}
  \begin{tabular*}{\textwidth} { |p{3cm}L{3cm}L{2.6cm}L{2.6cm}L{3.2cm}|}
  \hline
  	 &      &   &  &  \\
  \textbf{type} & \textbf{method}  & \textbf{n=d} & \textbf{n=2d} & \textbf{n=3d}\\ 
   &      &   &  &  \\
   \hline
   \cellcolor{gray!20}&      &   &  &  \\
   \cellcolor{gray!20}&MixMPLN      &0.87 (14m) &0.92 (10.5m) &0.93( 13m)\\
   \textbf{K=1}\cellcolor{gray!20}&      &   &  &  \\
   \cellcolor{gray!20}&MixGGM   &0.87 (5m) &0.91 (3m) &0.94 (4m)\\
   \cellcolor{gray!20}&      &   &  &  \\
    \hline
    \cellcolor{gray!20}&      &   &  &  \\
   \cellcolor{gray!20}&MixMPLN      &0.70 (48m) &0.76 (42m) &0.80 (3h)\\
   \textbf{K=2}\cellcolor{gray!20}&      &   &  &  \\
   \cellcolor{gray!20}&MixGGM  &\textbf{0.72} (36m) &\textbf{0.79} (33m) &\textbf{0.84} (28m)\\
   \cellcolor{gray!20}&      &   &  &  \\
    \hline
    \cellcolor{gray!20}&      &   &  &  \\
   \cellcolor{gray!20}&MixMPLN      &0.64 (1.1h) &0.70 (45m) &0.75 (2.3h)\\
   \textbf{K=3}\cellcolor{gray!20}&      &   &  &  \\
   \cellcolor{gray!20}&MixGGM   &\textbf{0.67} (1.5h) &\textbf{0.74} (1.4h) &\textbf{0.81} (1.3h)\\
   \cellcolor{gray!20}&      &   &  &  \\
   
    \hline
    \end{tabular*}
 	\end{center}
\end{table}  
\begin{table}[htp]
\caption{AUC performance on synthetic data with "random" graph type with $d=127$. "$K$" indicates the number of components to generate the synthetic data. "$n$" is the number of samples in each component. \cite{mae} Running time is indicated next to the AUC while s, h and d are abbreviations for seconds, hours and days respectively.}
\begin{center}
   
  \def\arraystretch{.5}
\begin{tabular*}{\textwidth} { |p{3cm}L{3cm}L{2.6cm}L{2.6cm}L{3.2cm}|}  \hline
  	&      &   &  &  \\
  \textbf{type} & \textbf{method}  & \textbf{n=d} & \textbf{n=2d} & \textbf{n=3d}\\ 
  &      &   &  &  \\
   \hline
   \cellcolor{gray!20}&      &   &  &  \\
   \cellcolor{gray!20}&MixMPLN      &0.57 (6m) &0.62 (10m) &0.65 (6m)\\
   \textbf{K=1}\cellcolor{gray!20}&      &   &  &  \\
   \cellcolor{gray!20}&MixGGM   &0.56 (5m) &0.62 (4m) &0.66 (4m)\\
   \cellcolor{gray!20}&      &   &  &  \\
    \hline
    \cellcolor{gray!20}&      &   &  &  \\
   \cellcolor{gray!20}&MixMPLN      &0.53 (1.2h) &0.55 (45m) &0.55 (2.1h)\\
   \textbf{K=2}\cellcolor{gray!20}&      &   &  &  \\
   \cellcolor{gray!20}&MixGGM  &0.52 (32m) &0.55 (28m) &\textbf{0.57} (28m)\\
   \cellcolor{gray!20}&      &   &  &  \\
    \hline
    \cellcolor{gray!20}&      &   &  &  \\
   \cellcolor{gray!20}&MixMPLN      &0.51 (1h) &0.53 (1h) &0.53 (3h)\\
   \textbf{K=3}\cellcolor{gray!20}&      &   &  &  \\
   \cellcolor{gray!20}&MixGGM   &0.51 (1h) &0.53 (1h) &\textbf{0.55} (1h)\\
   \cellcolor{gray!20}&      &   &  &  \\
   
    \hline
    \end{tabular*}
 	\end{center}
\end{table}  
\begin{table}[htp]
\caption{AUC performance on synthetic data with "scale-free" graph type with $d=127$. "$K$" indicates the number of components to generate the synthetic data. "$n$" is the number of samples in each component. Running time is indicated next to the AUC while s, h and d are abbreviations for seconds, hours and days respectively.}
\begin{center}
   
  \def\arraystretch{.5}
\begin{tabular*}{\textwidth} { |p{3cm}L{3cm}L{2.6cm}L{2.6cm}L{3.2cm}|}  \hline
  	&      &   &  &  \\
  \textbf{type} & \textbf{method}  & \textbf{n=d} & \textbf{n=2d} & \textbf{n=3d}\\ 
  &      &   &  &  \\
   \hline
   \cellcolor{gray!20}&      &   &  &  \\
   \cellcolor{gray!20}&MixMPLN      &0.64 (6m) &0.75 (6m) &0.83 (10m)\\
   \textbf{K=1}\cellcolor{gray!20}&      &   &  &  \\
   \cellcolor{gray!20}&MixGGM   &\textbf{0.66} (5m) &\textbf{0.77} (4m) &\textbf{0.85} (4m)\\
   \cellcolor{gray!20}&      &   &  &  \\
    \hline
    \cellcolor{gray!20}&      &   &  &  \\
   \cellcolor{gray!20}&MixMPLN      &0.58 (28m) &0.64 (42m) &0.63 (2h)\\
   \textbf{K=2}\cellcolor{gray!20}&      &   &  &  \\
   \cellcolor{gray!20}&MixGGM  &0.59 (39m) &0.65 (37m) &0.64 (40m)\\
   \cellcolor{gray!20}&      &   &  &  \\
    \hline
    \cellcolor{gray!20}&      &   &  &  \\
   \cellcolor{gray!20}&MixMPLN      &0.53 (1h) &0.57 (2.2h) &0.57 (4h)\\
   \textbf{K=3}\cellcolor{gray!20}&      &   &  &  \\
   \cellcolor{gray!20}&MixGGM   &0.54 (2h) &0.58 (1.4h) &\textbf{0.60} (56m)\\
   \cellcolor{gray!20}&      &   &  &  \\
    \hline
    \end{tabular*}
 	\end{center}
\end{table}  
We used the synthetic data generated from the different graph types for various values of $K, d,$ and $n$ to evaluate the methods with sparsity constraints . For each dataset, every method was run three times with different random starting points, and the solution with the largest log-likelihood value was selected for that method. All methods were assessed with respect to their ability to recover the original graph topology, and thus, the area under the ROC curve (AUC) was used as the performance evaluation criterion. In addition, we also evaluated the running times of the methods. Experiments were carried out using an Intel(R) Xeon(R) E5-2640 v4@2.40GHZ machine.

The performance results on band, cluster, random and scale-free graphs are given in Tables 5.1, 5.2, 5.3 and 5.4 respectively. In the tables, the best AUC value (with at-least $2\%$ difference) in each experiment is shown in bold. Several patterns are noteworthy: (a) For each graph type and each $K$, as expected, the AUC value for a method increases with increasing $n$; this is a reflection of improved inference capability of each method with increasing sizes of the input data. (b) The ability of a method to recover the original graph is a function of the graph type. In general, the methods tend to be able to recover the topologies for band graphs and cluster graphs more accurately in comparison to topologies for scale-free graphs and random graphs (this is reflected in the comparatively higher AUC values for band and cluster graphs for each method). (c) MixGGM, which uses log-ratio transformation to deal with the compositionality of the data, generally tends to outperform MixMPLN on the different graph types. 
\begin{itemize}
    \item (Band graph) When $K=1$, MixMPLN and MixGGM have comparable performances. For $K>1$ MixGGM performs slightly better than MixMPLN (except for $K=3, n=3d$). (Table 5.1) 
    \item (Cluster graph) When $K=1$, the result of the MixMPLN and MixGGM are comparable. For other cases MixGGM performs better. The maximum difference between the result of the two methods is when $K=3$ and $n=3d$ ($81\%$ vs $75\%$). (Table 5.2)
   \item (Random graph) MixGGM performs better when $K>1$ and $n=3d$ (although the difference range is $2\%$ ). For the other cases, the two methods are comparable. (Table 5.3) 
   \item (Scale-free graph) When $K=1$ and any value of $n$, MixGGM performs better. The other case which MixGGM works better is $K=3$ and $n=3d$ where its AUC value is $3\%$ higher than MixMPLN result. (Table 5.4) 
\end{itemize}
 (d) In addition, the average runtime of MixGGM is less than that of MixMPLN.

\section{Results on real data}

We applied MixGGM and MixMPLN to a genus-level sample-taxa matrix from 16S data generated from stool samples (HMP \cite{HMP}). The data consists of taxa counts derived from 319 stool samples and 95 genera. We set the number of components $K = 1$. Each method was run three times with different random starting points, and the model with the largest log-likelihood value was selected for that method. While we do not know the ground-truth network(s) for this dataset, it is possible to evaluate the consistency of the results generated by the two methods. Figure 5.1 depicts the scatter plot of the upper triangle entries of the estimated partial correlation matrices produced by MixMPLN and MixGGM. We observe that there is good correlation between the two methods, with respect to the inference of pairwise associations and their strengths (i.e. edge weights). However, there are low weight edges in the model inferred by one method that are absent (i.e. assigned a weight of 0) in the model inferred by the other method. These are likely noise edges, and need to be explored further.       

\begin{figure}
    \centering
    \includegraphics[width=\linewidth]{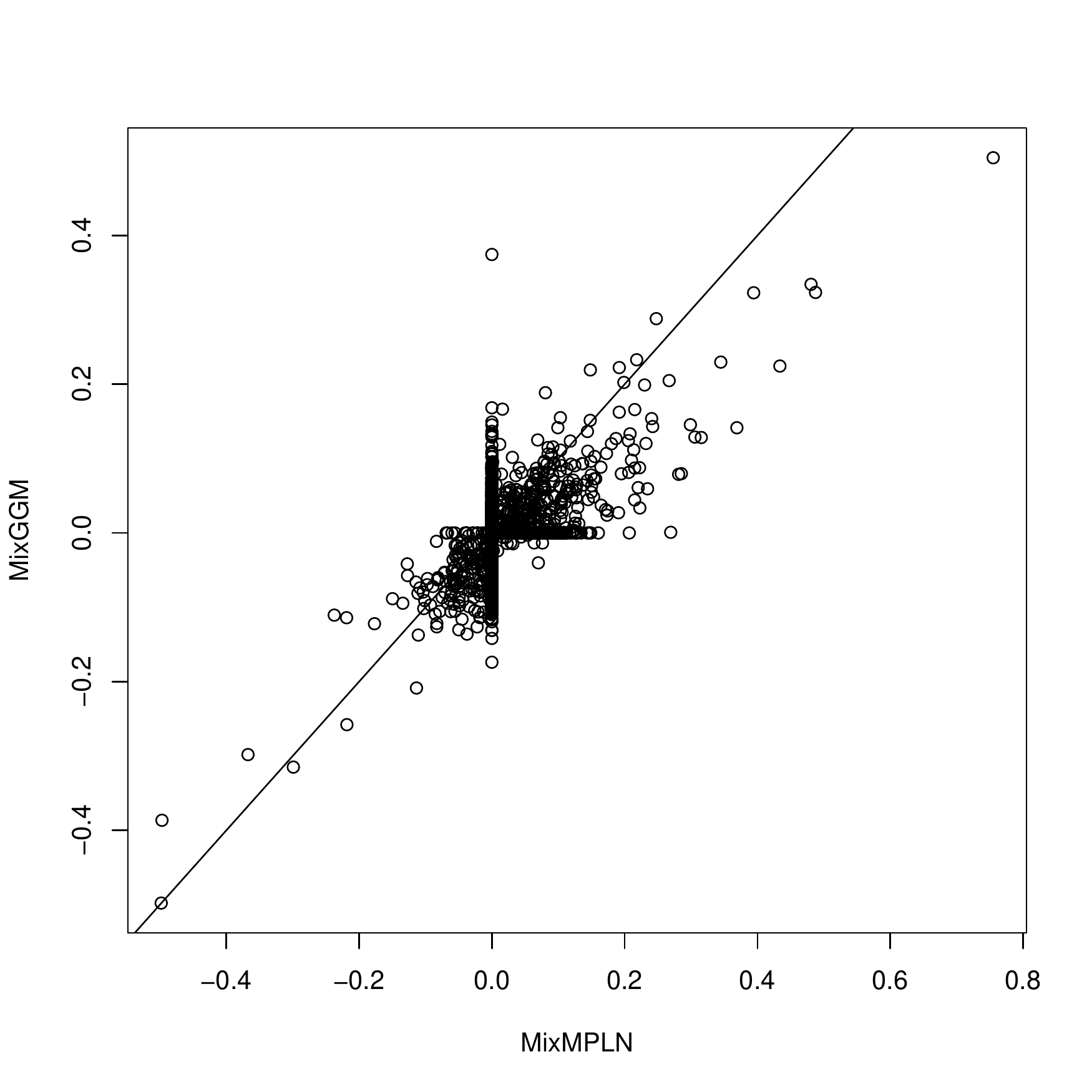}
    \caption{Scatterplot showing the entries of the partial correlation matrices generated by MixMPLN and MixGGM on the real dataset, with K=1. The line $y = x$ is also shown.}
\end{figure}  
 
\section{Summary}
In this chapter, we introduced MixGGM, to infer multiple networks from microbial count data. We used synthetic datasets to evaluate these approaches with respect to recovering the original graph topologies, for different types of graphs. While the AUC values increase with increasing sample size of the input data, we observe that this is a function of the graph type. The methods tend to be able to recover the topologies for band graphs and cluster graphs more accurately in comparison to topologies for scale-free graphs and random graphs. Another observation is that MixGGM (which uses log-ratio transformation) generally tends to perform better compared to MixMPLN. However, the MPLN distribution, which forms the basis of the MixMPLN framework, allows for the inclusion of biological and/or technical co-variates in the Poisson layer to control for confounding factors \cite{Ma2008 , Biswas}. \\
In the next chapter (chapter 6), we explores methods to estimate the number of components in a mixture model, and will introduce a novel and practical framework for model selection issue in modeling microbial data in a mixture framework.






\chapter{MODEL SELECTION}
So far, in all the proposed algorithms, the number of components($K$) was assumed to be known. This chapter's content is to find a framework to estimate the best number of components in modeling a sample-taxa matrix as a mixture model. As we are not aware of the original distribution which generated the data, we do not know the ground truth of our component generators and a technique is needed to find the best number of components. The best number of the components will be $K$ which is considered as the input of the previous introduced algorithms to infer the microbial interactions.\\

There are many methods to find the optimal number of components some of which are not accurate enough or computationally expensive. One simple approach is to consider non-parametric methods for selecting the number of components such as calculating the Silhouete score \cite{silhouete}. This score checks how much the clusters are compact and well separated by calculating the mean distance between a sample and all other points in the same cluster and the mean distance between a sample and all other points in the next nearest cluster. The Silhouete score will be calculated for different number of components and the number of components with the higher Silhouete score is selected as the optimal one.\\

Another criterion to detect the best number of components in a mixture model comes from a Bayesian framework for model selection. The model should be run for different values for number of components and best model will be detected based on the decision criteria. This model selection criteria is trading off between the log-likelihood and model complexity in order to avoid over-fitting. Among these types of methods we note here the Bayesian Information Criterion (BIC) \cite{BIC}. The Bayesian information criterion (BIC) is define as:\\
\\
$BIC = -log(L({\theta}|data)) + Klog(n)$
\\
Where $L$ is the likelihood function of the $\theta$ parameters given data. $k$ is number of components and $n$ is number of parameters in the model. \\
In finding the best number of components using BIC method, the best model will be the one that has the lowest BIC. \\

Reversible Jump MCMC (RJMCMC)\cite{Green1995} is another method which gives us the opportunity to choose from a collection of models, but it is computationally intense. Unlike other sampling algorithms, e.g. Gibbs sampling\cite{cn} or the generic Metropolis-Hastings algorithms\cite{dl}, this method explores different submodels and bridges between them by jumping across spaces of varying dimension to find the best generalized fit to the data.\\ 

All the methods which are mentioned so far need to fit the data to the model several times using different number of components to find the best model. Thus, finding the optimal number of the component associated with the data can be computationally expensive. This fact is the reason that we are interested in the Variational Inference(VI) \cite{VI} algorithm which can be used to estimate the number of components by running the algorithm on the data once instead of running it several times for different models.\\

VI optimizes the marginal log-likelihood function associated with the data instead of the regular log-likelihood function and provides approximate solutions to the inference problem. It gives us the opportunity to consider a prior distribution for a parameter in the model. In other words, each point parameter (called hidden variable) will be replaced with a distribution (called variational posterior). Optimizing the marginal log-likelihood function, tries to make the variational posteriors as close as possible to the true posteriors. This approach runs the mixture model on a large fixed number of components when the only parameters are mixing coefficients and everything else is a hidden variable. Therefore, mixing coefficients corresponding to unwanted components will go to zero after the convergence\cite{corduneanu}.     \\

In the following equations, $Q(\theta)$ includes all the posteriors for the hidden variables of $\theta$ and only $\pi$ is the parameter including the mixing coefficients to be optimized. Following equations \cite {VI,ml} show how we calculate new objective function in VI.
\begin{equation}
ln(P(D|\pi ))=ln\int P(D,\theta |\pi)d\theta 
\end{equation}
Where $D$ is data, $\theta$ includes all the hidden variables and $\pi$ is the parameter which includes all the mixing coefficients. \\
Equation 6.1 can be written as following equation: 
\begin{equation}
=ln \int  (Q(\theta)\frac{P(D,\theta |\pi)}{Q(\theta)})d\theta  
\end{equation}
Where the lower bound for equation 6.2 can be written as: 
\begin{equation}
\geq \int Q(\theta) ln\frac{ P(D,\theta |\pi)}{Q(\theta)}d\theta  
\end{equation}
Based on previous equation, we can define the Evidence Lower Bound (Elbo) function which is our marginal log-likelihood\cite {VI}. 
\begin{equation}
Elbo(\theta) = \int Q(\theta) ln\frac{ P(D,\theta |\pi)}{Q(\theta)})d\theta  
\end{equation}
\\
In the previous equation, Elbo function which is the lower bound for the marginal log-likelihood function is shown. The Elbo function is our objective function to maximise in VI for model selection. \\

One way to solve the explained optimization problem is to find the updating formulas for the unknown parameters. By considering a specific situation that general posterior distribution has a conjugate form which is indicated in equation(6.5), the optimal posterior distributions are then given by equation(6.6)\cite {VI} \cite{sanaz1} \cite{sanaz2}:

\begin{equation}
    Q(\theta) = \prod Q_{i}(\theta_{i})
\end{equation}
optimal posterior for each parameter is calculated as\cite {VI} :
\begin{equation}
    Q_{i}(\theta_{i}) = \frac{exp<ln P(D,\theta)>_{k\neq i}}{\int exp<ln P(D,\theta)>_{k\neq i}d\theta_{i}}
\end{equation}

Knowing the following fact, we can easily obtain the updating formulas for finding the optimal parameters. \\

Another way to solve the variational Inference is to use Variational Auto Encoder(VAE)(depicted in Figure 6.1). VAE is used to reconstruct the input data, by setting the loss to minimise the distance between the input and reconstructed data in output layer. By defining a proper loss and considering the KL divergence between the prior and posterior distribution of the parameters we can get to the optimal variables for the variational inference problem. \\

\begin{figure}
    \centering
\includegraphics[width=\textwidth]{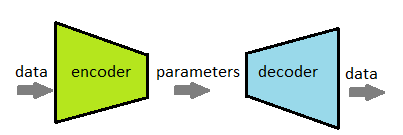}
 \caption{Variational Auto Encoder's structure}
 \end{figure}

To find a proper elbo function for VAE in solving the VI, we can write the equation 6.4 as following equation, where $E$ depicts the expectation function\cite{Hoff,nn}:\\
\begin{equation}
    Elbo(Q) = E[log P(D,\theta)] - E[log Q(\theta)] 
\end{equation}
Which can be written as\cite{Hoff}:
\begin{equation}
    Elbo(Q) = E[logP(\theta)] + E[logP(D|\theta)] - E[log Q(\theta)] 
\end{equation}
Knowing the definition of KL divergence, we can get to a proper Elbo function for VAE in modeling the VI\cite{Hoff}:
\begin{equation}
    Elbo(Q) = E[logP(D|\theta)] - KL(Q(\theta)||P(\theta))
\end{equation}
\\
In the next section, we applied the tensorflow implementation for the Gaussian mixture model and extended it to the Poisson log normal distribution. \\

\section{Implementing the VI for model selection in tensorflow framework }
\subsection{Mixture of Gaussians }
We have used the tensorflow framework to optimize the mentioned Elbo function by setting proper prior distribution for mixture of Gaussian distribution:\\
For the Gaussian mixture model the prior distributions for the parameters are described by the following equations: \\
$X_{i} \sim Normal(\mu_{k} , \delta _{k})$ \\
Where $k$ is number of components, $\mu_{k} , \delta _{k}$ are mean and standard deviation corresponding to the $k_{th}$ component. \\
The only parameters in the model are the mixing coefficients which are forced to sum up to one. 
Following equations show the prior distribution that we selected for the mean and the inverse of variances: \\
$\delta_{k,d}^{-2} \sim Gamma(\alpha_{k,d} , \beta_{k,d})$\\
$\mu_{k,d} \sim Normal(l_{k,d} , s_{k,d})$\\

In the preprocessing stage, Principle Component Analysis(PCA) is applied to the normalized data. We selected number of components by setting $90\%$ cut-off threshold on cumulative variance. After PCA another normalization is applied to the data to make the convergence in the optimization process faster.   

One main disadvantage of using tensorflow in solving the VI for model selection is that for any new dataset, all the hyper-parameters of the optimization should be specialized in order to get to the steady state level in the loss function value. These hyper-parameters include learning rate, number of iterations, batch size and initial points for parameters inside the distributions. However we can get to the optimal number of components in this method, the mentioned disadvantage makes it impractical.    \\
Figure 6.1 shows the result of applying the variational inference method implemented in tensorflow for data from mixture of two components which each component follows the Gaussian probability distribution. Figure 6.2 shows the result of applying the variational inference method implemented in tensorflow for data from mixture of three components which each component follows the Gaussian probability distribution.\\

\begin{figure}[h!]
\centering  
\subfigure[]{\includegraphics[width=0.49\linewidth]{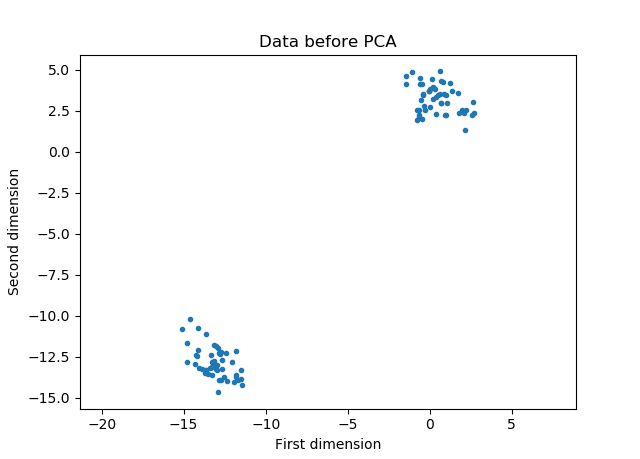}}
\subfigure[]{\includegraphics[width=0.49\linewidth]{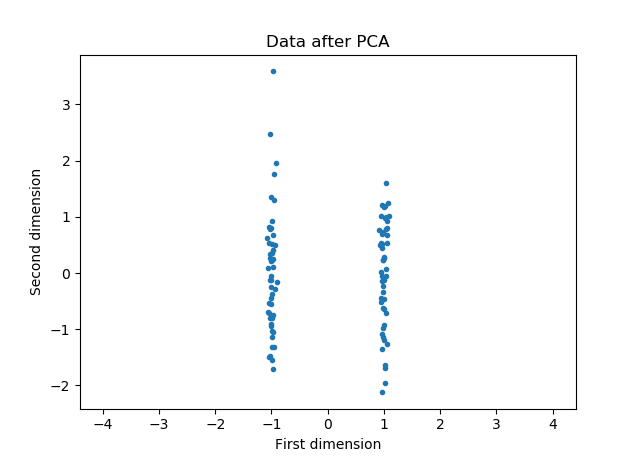}}
\subfigure[]{\includegraphics[width=0.49\linewidth]{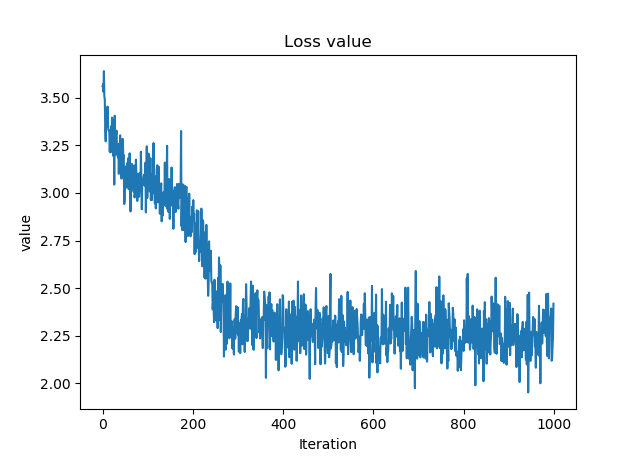}}
\subfigure[]{\includegraphics[width=0.49\linewidth]{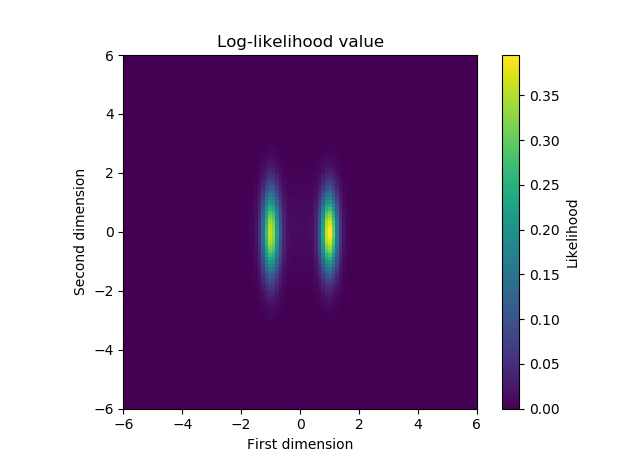}}
\caption{The result of applying the variational inference method implemented in tensorflow one a data coming from mixture of two components which each component follows the Gaussian probability distribution. Initial values for the mixing coefficients were .2 for each components where the number of components are 5 and the algorithm converges to the [0.03 , 0.46 , 0.02 , 0.45 , 0.04] as mixing coefficients which shows that the best number of components is two components as three other mixing coefficients are decreasing and two other are increasing while they all started with the same value of 0.2.\textbf{(a)}First two dimensions of the data for a data with 100 samples and 100 dimension which 50 samples are coming from the first component 50 samples are coming from the second component. \textbf{(b)} Selected dimensions after applying the PCA  \textbf{(c)} Loss function value which shows the convergence of the algorithm as loss value reaches to the steady state. Hyper parameters to reach this convergence are: number of epochs =1000, learning rate = 0.005 , batch size = 50  \textbf{(d)}Log likelihood value which shows two selected components  }
\end{figure}

\begin{figure}[h!]
\centering  
\subfigure[]{\includegraphics[width=0.49\linewidth]{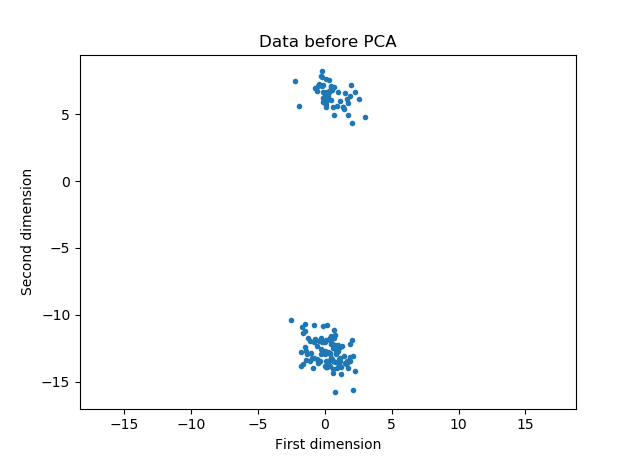}}
\subfigure[]{\includegraphics[width=0.49\linewidth]{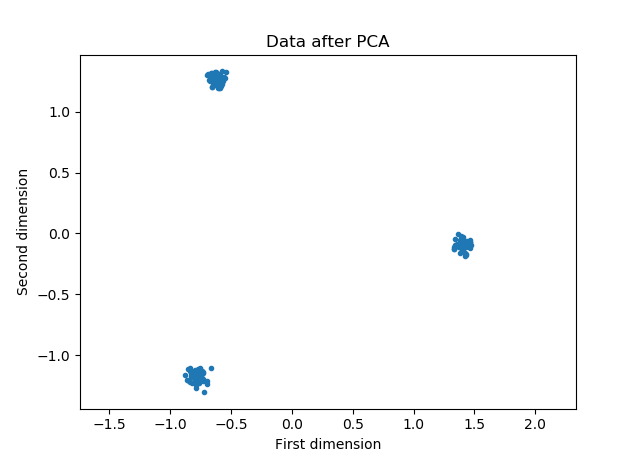}}
\subfigure[]{\includegraphics[width=0.49\linewidth]{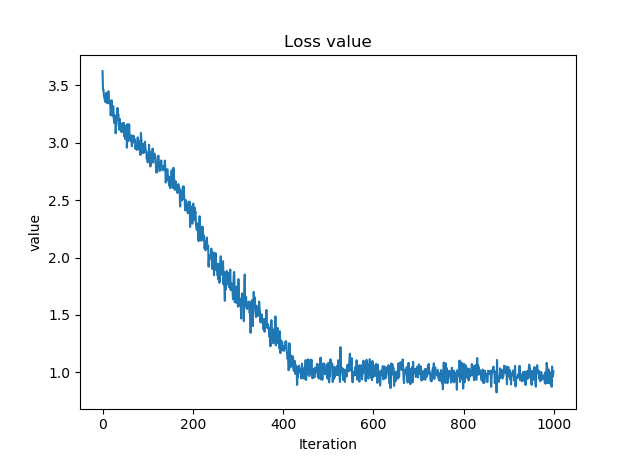}}
\subfigure[]{\includegraphics[width=0.49\linewidth]{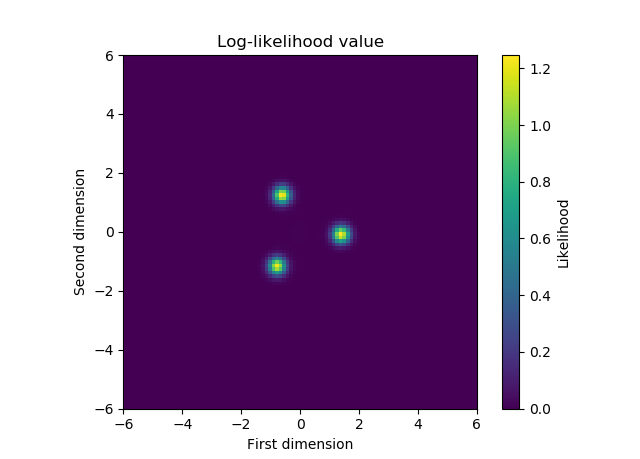}}
\caption{The result of applying the variational inference method implemented in tensorflow one a data coming from mixture of two components which each component follows the Gaussian probability distribution. Initial values for the mixing coefficients were .2 for each components where the number of components are 5 and the algorithm converges to the [0.01 , 0.32 , 0.02 , 0.31 , 0.24] as mixing coefficients which shows that the best number of components is two components as three other mixing coefficients are decreasing and two other are increasing while they all started with the same value of 0.2.\textbf{(a)}First two dimensions of the data for a data with 100 samples and 100 dimension which 50 samples are coming from the first component 50 samples are coming from the second component. \textbf{(b)} Selected dimensions after applying the PCA  \textbf{(c)} Loss function value which shows the convergence of the algorithm as loss value reaches to the steady state. Hyper parameters to reach this convergence are: number of epochs =1000, learning rate = 0.005 , batch size = 50  \textbf{(d)}Log likelihood value which shows two selected components }
\end{figure}

\subsection{Mixture of Poisson Log Normal}

In the next step, we extended our implementation for the variational inference in tensorflow for mixture of poisson log normal distribution by setting the proper distributions as following equations: \\
$X_{i} \sim Poisson(e^{\lambda_{i}})$\\
$\lambda_{i} \sim Normal(\mu_{k} , \delta _{k})$\\
Here data for each component is coming from a Poisson Log-Normal distribution, where $\lambda$ is latent variable , $i$ number of sample and $k$ number of component. \\
The only parameters in the model are the mixing coefficients which are forced to sum up to one.\\
Prior for hidden variables are defines the same as previous simulation:\\
$\delta_{k,d}^{-2} \sim Gamma(\alpha_{k,d} , \beta_{k,d})$\\
$\mu_{k,d} \sim Normal(l_{k,d} , s_{k,d})$\\
Figure 6.4 depicts two plots corresponding to the loss function in our optimization problem for two different data set. Reaching the loss function value to a steady state guarantees the convergence of the optimization problem for these datasets. 
\begin{figure}
\centering  
 \subfigure[]{\includegraphics[width=0.45\linewidth]{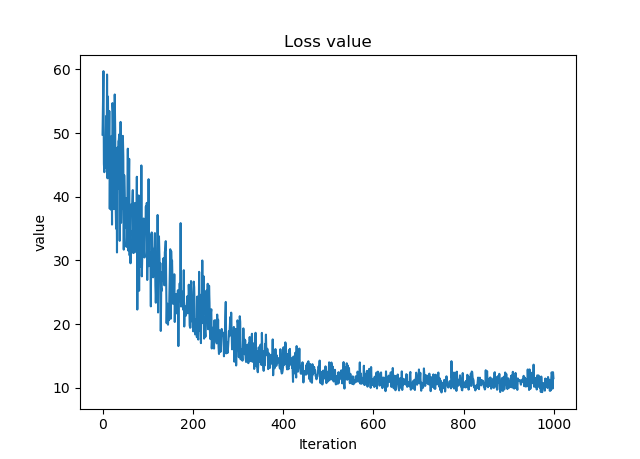}}
\subfigure[]{\includegraphics[width=0.45\linewidth]{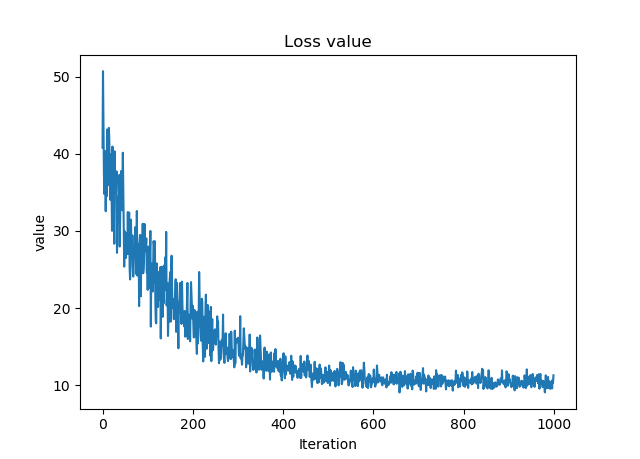}}
\caption{(a)Loss function value related to a data coming from two components with 100 samples and 100 dimensions. The final mixing coefficients values are: [0.05, 0.46, 0.35, 0.03, 0.11] (b) Loss function value related to a data coming from three components with 150 samples and 100 dimensions. The final mixing coefficients values are:[0.07, 0.36, 0.28, 0.06, 0.23]}
\end{figure}\\

As it is shown in Figure 6.1, Figure 6.2 and Figure 6.3, if we get to the steady state of the loss function, we will find the optimal number of the components. The issue is tuning the hyper-parameters for each new input data to get to the steady state in the loss function. This issue makes the VI with tensorflow an impractical framework and persuades us to find a practical framework to solve the VI for model selection. \\
The mentioned reason motivated us to derive the formula directly as will be explained in the next sections.\\
\\
\section{ VI formulation for GMM}

As it is mentioned in the previous section, using tensorflow to solve the VI for model selection is not practical. This issue motivates us to find a closed form solution for the problem. We have included the formula derivation for the model selection using the Variational Inference in Gaussian mixture model in this section\cite{corduneanu}. As we discussed previously to estimate the number of components using the VI, we optimise the marginal log-likelihood instead of regular log-likelihood function. Here we consider the posterior distribution of variables as multiplication of posterior for each variable, thus we are able to use equation 6.6 to find the optimal posterior functions\cite{corduneanu}.   \\
\begin{equation}
    P(D,\mu,T,S) = P(D|\mu,T,S)P(S|\pi)P(\mu))P(T)
\end{equation}
Where $D$ is data, $\mu,T$ are mean, precision matrices for all components. $S$ is latent variable. $s_{in} =1 $ when $i_{th}$ sample is coming from $n_{th}$ component, and $s_{in} =0 $ vice versa. $\pi$ is the only parameter in the model which includes mixing coefficients.  
\begin{equation}
    P(D|\pi,\mu,T) = \prod_{n=1}^{N} [\sum_{i=1}^{M}\pi_{i}N(x_{n}|\mu_{i},T_{i})]
\end{equation}
Where $N$ is number of samples, $M$ is number of components. Using latent variable $S$, we can re write the equation 6.11 as following equation: 
\begin{equation}
    P(D|\mu,T,S) \sim \prod_{i=1}^{M}\prod_{n=1}^{N}   (N(\mu_{i},T_{i})^{S_{in}})
\end{equation}
Prior distributions are shown in following equations. Drichlete distribution for $S$, Normal distribution for mean vectors and Wishart distribution for Precision matrices : 
\begin{equation}
    P(S|\pi) \sim  \prod_{i=1}^{M}\prod_{n=1}^{N} (\pi_{in})^{S_{in}}
\end{equation}
\begin{equation}
    P(\mu) \sim  \prod_{i=1}^{M}  N(0,\beta I)
\end{equation}
\begin{equation}
    P(T) \sim  \prod_{i=1}^{M}  W(\nu , V)
\end{equation}

Next equations shows posterior distributions for hidden variables: 
\begin{equation}
    Q(\mu , T , S)= Q(\mu)Q(T)Q(S)
\end{equation}
\begin{equation}
    Q(\mu) \sim \prod_{i=1}^{M} N(m_{\mu}^i, T_{\mu}^i)
\end{equation}
\begin{equation}
    Q(T) \sim \prod_{i=1}^{M} W(\nu_{T}^i, V_{T}^i)
\end{equation}
\begin{equation}
    Q(S) \sim \prod_{n=1}^{N}\prod_{i=1}^{M} p_{in}^{S_{in}}
\end{equation}

\textbf{E-step to update the $p_{in}$----------------------------------------------------------------------------------------}\\

$lnQ^{*}(S) = E_{\mu , T}[lnP(D,\mu,T,S|\pi)] + constant $\\
$lnQ^{*}(S) = E_{\mu , T}[lnP(D|\mu , T, S) + lnP(S|\pi) + lnP(\mu) + lnP(T)] + constant$\\
$lnQ^{*}(S) = E_{\mu , T}[lnP(D|\mu , T, S) + lnP(S|\pi)] + constant$\\
$\sum_{i=1}^{M}\sum_{n=1}^{N} S_{in} P_{in} = E_{\mu,T}[\sum_{i=1}^{M}\sum_{n=1}^{N} S_{in}( \frac{ln(|T_{i}|)}{2}-1/2(x_{n} - \mu_{i})^{t}T_{i}(x_{n} - \mu_{i})) +\sum_{i=1}^{M}\sum_{n=1}^{N}   S_{in} \pi_{i}]+ constant$\\
$\sum_{i=1}^{M}\sum_{n=1}^{N} S_{in} P_{in} = E_{\mu,T}[\sum_{i=1}^{M}\sum_{n=1}^{N} S_{in}( \frac{ln(|T_{i}|)}{2}-1/2Tr(T_{i}(x_{n} - \mu_{i})^{t}(x_{n} - \mu_{i}))) + \sum_{i=1}^{M}\sum_{n=1}^{N} S_{in} ln\pi_{i}]+ constant$\\

\begin{equation}
P_{in} = \frac{\tilde{P_{in}}}{\sum_{j=1}^{M}\tilde{P_{in}}}
\end{equation}

\begin{equation}
\tilde{P_{in}} = exp(\frac{< ln|T_{i}|>}{2}  + ln\pi _{i} - 1/2Trace(<|T_{i}|>(x_{n}x_{n}^{T} -<\mu_{i}>x_{n}^{T} - x_{n}<\mu_{i}>^{T} + <\mu_{i}\mu_{i}^{T}>))  
\end{equation}

\textbf{E-step to update the $T_{\mu}^{i}$ and $m_{\mu}^{i}$ ------------------------------------------------------------------------------}\\

$lnQ^{*}(\mu) = E_{T,S}[lnP(D,\mu,T,S|\pi)] + constant $\\
$lnQ^{*}(\mu) = E_{T,S}[lnP(D|\mu , T, S) + lnP(S|\pi) + lnP(\mu) + lnP(T)] + constant$\\
$lnQ^{*}(\mu) = E_{T,S}[lnP(D|\mu , T, S) + lnP(\mu) + constant$\\
$\sum_{i=1}^{M} ln N(m^{i}_{\mu} , T^{i}_{\mu})= E_{T,S}[\sum_{i=1}^{M}\sum_{n=1}^{N} S_{in}( \frac{ln(|T_{i}|)}{2}-1/2(x_{n} - \mu_{i})^{t}T_{i}(x_{n} - \mu_{i})) \\
+  \sum_{i=1}^{M} lnN(0 , \beta I)]+ constant$\\
$\sum_{i=1}^{M} -1/2(\mu_{i} - m^{i}_{\mu})^{t}T^{i}_{\mu}(\mu_{i} - m^{i}_{\mu}) = E_{T,S}[\sum_{i=1}^{M}\sum_{n=1}^{N} S_{in}(-1/2(x_{n} - \mu_{i})^{t}T_{i}(x_{n} - \mu_{i}) )+\\ \sum_{i=1}^{M}(-1/2(\mu_{i})^{t}(\beta I)(\mu_{i}))]+ constant$

\begin{equation}
T_{\mu}^{i} = \beta I + <T_{i}>\sum_{n=1}^{N}<S_{in}>
\end{equation}

\begin{equation}
m_{\mu}^{i} = T_{\mu}^{i^{-1}}<T_{i}>\sum_{n=1}^{N}x_{n}<S_{in}>
\end{equation}

\textbf{E-step to update the $\nu_{T}^{i}$ and $V_{T}^{i}$ -----------------------------------------------------------------------------}\\

$lnQ^{*}(T) = E_{\mu,S}[lnP(D,\mu,T,S|\pi)] + constant $\\
$lnQ^{*}(T) = E_{\mu,S}[lnP(D|\mu , T, S) + lnP(S|\pi) + lnP(\mu) + lnP(T)] + constant$\\
$lnQ^{*}(T) = E_{\mu,S}[lnP(D|\mu , T, S) + lnP(T) + constant$\\
$\sum_{i=1}^{M} ln W(\nu^{i}_{T} , V^{i}_{T})= E_{\mu,S}[\sum_{i=1}^{M} \sum_{n=1}^{N} S_{in}( \frac{ln(|T_{i}|)}{2}-1/2(x_{n} - \mu_{i})^{t}T_{i}(x_{n} - \mu_{i})) +  \sum_{i=1}^{M} ln W(\nu , V)]+ constant\\$
$\sum_{i=1}^{M} ln B(\nu_{T}^{i} ,V_{T}^{i} ) + 1/2(\nu_{T}^{i} - d -1)ln|T_{i}|-1/2Tr((V^{i}_{T})^{-1}T_{i}) =E_{\mu,S}[\sum_{i=1}^{M} \sum_{n=1}^{N} S_{in}(-1/2Tr((x_{n} - \mu_{i})^{t}(x_{n} - \mu_{i})T_{i}) ) + \sum_{i=1}^{M}(ln B(\nu ,V) + 1/2(\nu - d -1)ln|T_{i}|-1/2Tr(V^{-1}T_{i}))]+ constant$
\begin{equation}
\nu _{T}^{i} = \nu + \sum_{n=1}^{N}<S_{in}>
\end{equation}

\begin{equation}
\begin{split}
V_{T}^i = V + \sum_{n=1}^{N}x_{n}x_{n}^{T}<S_{in}> - \sum_{n=1}^{N}x_{n}<S_{in}><\mu_{i}^{T}> - \\<\mu_{i}>\sum_{n=1}^{N}x_{n}^{T}<S_{in}> +<\mu_{i}\mu_{i}^{T}>\sum_{n=1}^{N} <S_{in}>
\end{split}
\end{equation}

\textbf{calculating the expectations------------------------------------------------------------------------------------}\\

\begin{equation}
\begin{split}
<S_{in}>=P_{in}\\
<\mu_{i}>=m_{\mu}^{i}\\
<\mu_{i} \mu_{i}^{T}>= T_{\mu}^{i^{-1}} + m_{\mu}^{i}m_{\mu}^{i^{T}}\\
<T_{i}>= \nu _{T}^{i} V_{T}^{i^{-1}}\\
<ln|T_{i}|>=\sum_{s=1}^{d}\Psi(\frac{\nu_{T}^{i} + 1 - s}{2}) +dln2 - ln|V_{T}^{i}| \\ 
\end{split}
\end{equation}

\textbf{calculating the Elbo function---------------------------------------------------------------------------------}\\

\begin{equation}
\begin{split}
L = <lnP(D|\mu , T , S))> + <lnP(S)> + <lnP(\mu)>+ <lnP(T)>\\ - <lnQ(S)> - <lnQ(\mu)> - <lnQ(T)>
\end{split}
\end{equation}

\begin{equation}
\begin{split}
 <lnP(D|\mu , T , S)> = \sum_{i=1}^{M}\sum_{n=1}^{N}<S_{in}> (1/2<ln|T_{i}|> - d/2ln(2\pi)\\
- 1/2Tr(<T_{i}> ( x_{n}x_{n}^{T}-x_{n}<\mu_{i}>^{T} - <\mu_{i}>x_{n}^{T}+<\mu_{i}\mu_{i}^{T}>))))
\end{split}
\end{equation}

\begin{equation}
     <lnP(S)> = \sum_{i=1}^{M}\sum_{n=1}^{N}<S_{in}>ln\pi_{i}
\end{equation}

\begin{equation}
    <ln P(\mu)> = 2\frac{Md}{2}ln\frac{\beta }{2\pi} - \frac{\beta }{2}\sum_{i=1}^{M}<\mu_{i}^{t}\mu_{i}>
\end{equation}
\begin{equation}
\begin{split}
 <lnP(T)> =  M \left \{  -\frac{\nu d}{2}ln2 - \frac{d(d-1)}{4}ln\pi -\sum_{s=1}^{d}ln\Gamma (\frac{\nu + 1 - s}{2}) + (\nu)/2\ln|V|\right \} \\ + \frac{\nu - d -1}{2}\sum_{i=1}^{M}<ln|T_{i}|> - 1/2Tr(V\sum_{i=1}^{M}<T_{i}>)
\end{split}
\end{equation}

\begin{equation}
    <lnQ(S)>= \sum_{i=1}^{M}\sum_{n=1}^{N}<S_{in}>ln<S_{in}>
\end{equation}

\begin{equation}
    <lnQ(\mu)> = \sum_{i=1}^{M}\left \{ -d/2(1+ln2\pi) + 1/2ln|T^{i}_{\mu}|\right \} 
\end{equation}

\begin{equation}
    \begin{split}
        <lnQ(T)>= \sum_{i=1}^{M}( -\frac{\nu_{T}^{i}d}{2}ln2 -\frac{d(d-1)}{4}ln\pi -\sum_{i=1}^{d}ln\Gamma (\frac{\nu_{T}^{i} + 1 -s}{2}) \\+\nu_{T}^{i}/2ln|V_{T}^{i}| +\frac{\nu_{T}^{i} - d -1}{2}<ln|T_{i}|> -1/2Tr(V_{T}^{i}<T_{i}>) ) 
    \end{split}
\end{equation}

\textbf{M-step to update the mixing coefficients---------------------------------------------------------------}\\
In this step the mixing coefficient for each component is updated using the following equation:\\

$\pi_{i} = 1/N\sum_{n=1}^{N}p_{in}$

\section{ VI formulation for MPLN}
We have included the formula derivation for the model selection using the Variational Inference in mixture of Poisson log-Normal model in this section. As we discussed previously to estimate the number of components using the VI, we optimise the marginal log-likelihood instead of regular log-likelihood function. Here we consider the posterior distribution of variables as multiplication of posterior for each variable, thus we are able to use equation 6.6 to find the optimal posterior functions. In finding the updating equations for latent variables for $\lambda$, equation 6.6 will not extend to a closed form formula, therefor we optimise the Elbo function directly by finding its derivation respect to related parameters.  
\begin{equation}
    P(D,\lambda,\mu,T,S) = P(D|\lambda)P(\lambda|\mu,T,S)P(S|\pi)P(\mu)P(T)
\end{equation}
Where $D$ is data, $\mu,T$ are mean, precision matrices for all components.$\lambda$ and $S$ are latent variable. $s_{in} =1 $ when $i_{th}$ sample is coming from $n_{th}$ component, and $s_{in} =0 $ vice versa. $\pi$ is the only parameter in the model which includes mixing coefficients.
\begin{equation}
    P(D|\lambda) \sim Poisson(e^{\lambda}) 
\end{equation}
\begin{equation}
    P(\lambda|\pi,\mu,T) = \prod_{n=1}^{N} [\sum_{i=1}^{M}\pi_{i}N(\lambda_{n}|\mu_{i},T_{i})]
\end{equation}
Where $N$ is number of samples, $M$ is number of components. Using latent variable $S$, we can re write the equation 6.11 as following equation: 
\begin{equation}
    P(\lambda|\mu,T,S) \sim \prod_{i=1}^{M}\prod_{n=1}^{N}   (N(\mu_{i},T_{i})^{S_{in}})
\end{equation}

\begin{equation}
   P(D|\lambda)P(\lambda|\pi,\mu,T) = \prod_{n=1}^{N} [\sum_{i=1}^{M}\pi_{i}(\prod_{j=1}^{d}\frac{e^{-e^{\lambda^{i}_{nj}}}e^{\lambda^{i}_{nj}x_{nj}}}{x_{nj}!}\frac{\sqrt{det(T_{i})}}{(2\pi)^{d/2}}exp(-1/2(\lambda^{i}_{n} - \mu_{i})^{t}T_{i}(\lambda^{i}_{n} - \mu_{i})))
\end{equation}

Prior distributions are shown in following equations. Drichlete distribution for $S$, Normal distribution for mean vectors and Wishart distribution for Precision matrices :  

\begin{equation}
    P(S|\pi) \sim  \prod_{i=1}^{M}\prod_{n=1}^{N} (\pi_{in})^{S_{in}}
\end{equation}
 
\begin{equation}
    P(\mu) \sim  \prod_{i=1}^{M}  N(0,\beta I)
\end{equation}
\begin{equation}
    P(T) \sim  \prod_{i=1}^{M}  W(\nu , V)
\end{equation}

Next equations shows posterior distributions for hidden variables:

\begin{equation}
    Q(\lambda , \mu , T , S)= Q(\lambda)Q(\mu)Q(T)Q(S)
\end{equation}
\begin{equation}
    Q(\lambda) \sim \prod_{i=1}^{M}\prod_{n=1}^{N}\prod_{j=1}^{d} N(a_{nj}^i, b_{nj}^i)
\end{equation}
\begin{equation}
    Q(\mu) \sim \prod_{i=1}^{M} N(m_{\mu}^i, T_{\mu}^i)
\end{equation}
\begin{equation}
    Q(T) \sim \prod_{i=1}^{M} W(\nu_{T}^i, V_{T}^i)
\end{equation}
\begin{equation}
    Q(S) \sim \prod_{n=1}^{N}\prod_{i=1}^{M} p_{in}^{S_{in}}
\end{equation}

The difference between the formulation here and the GMM is that here in addition to previous hidden variables we have a latent variable of $\lambda$. Therefore we need to marginalize the equations based on the $\lambda$ too. \\
We are able to use equation 6.6 to derive closed form updating equations for all the parameters but $a$ and $b$ which are the parameters for posterior distribution of $\lambda$.\\
\textbf{E-step to update the $p_{in}$--------------------------------------------------------------------------------------}\\
\begin{equation}
lnQ^{*}(S) = E_{\lambda, \mu , T}[lnP(D,\lambda,\mu,T,S|\pi)] + constant 
\end{equation}

\begin{equation}
P_{in} = \frac{\tilde{P_{in}}}{\sum_{j=1}^{M}\tilde{P_{in}}}
\end{equation}

\begin{equation}
\begin{split}
\tilde{P_{in}} = exp(\sum_{j=1}^{d}(-<e^{\lambda_{nj}^{i}}>+ <\lambda_{nj}^{i}>x_{nj}  - logx_{nj}!) + \\ \frac{< ln|T_{i}|>}{2}  + ln\pi _{i} - 1/2Trace(<|T_{i}|>(<\lambda_{n}^{i}\lambda_{n}^{i^{T}}> -<\mu_{i}><\lambda_{n}^{i^{T}}>\\ - <\lambda_{n}^{i}><\mu_{i}>^{T} + <\mu_{i}\mu_{i}^{T}>)) 
\end{split}
\end{equation}

\textbf{E-step to update the $T_{\mu}^{i}$ and $m_{\mu}^{i}$ ----------------------------------------------------------------------------}\\
\begin{equation}
lnQ^{*}(\mu) = E_{\lambda,T,S}[lnP(D,\lambda,\mu,T,S|\pi)] + constant
\end{equation}

\begin{equation}
T_{\mu}^{i} = \beta I + <T_{i}>\sum_{n=1}^{N}<S_{in}>
\end{equation}

\begin{equation}
m_{\mu}^{i} = T_{\mu}^{i^{-1}}<T_{i}>\sum_{n=1}^{N}<\lambda_{n}><S_{in}>
\end{equation}

\textbf{E-step to update the $\nu_{T}^{i}$ and $V_{T}^{i}$ -----------------------------------------------------------------------------}\\
\begin{equation}
 lnQ^{*}(T) = E_{\lambda,\mu,S}[lnP(D,\lambda,\mu,T,S|\pi)] + constant   
\end{equation}

\begin{equation}
\nu _{T}^{i} = \nu + \sum_{n=1}^{N}<S_{in}>
\end{equation}

\begin{equation}
\begin{split}
V_{T}^i = V + \sum_{n=1}^{N}<\lambda_{n}\lambda_{n}^{T}><S_{in}> - \sum_{n=1}^{N}<\lambda_{n}><S_{in}><\mu_{i}^{T}> - \\<\mu_{i}>\sum_{n=1}^{N}<\lambda_{n}>^{T}<S_{in}> +<\mu_{i}\mu_{i}^{T}>\sum_{n=1}^{N} <S_{in}>
\end{split}
\end{equation}

\textbf{calculating the expectations------------------------------------------------------------------------------------}\\

\begin{equation}
\begin{split}
<S_{in}>=P_{in}\\
<\mu_{i}>=m_{\mu}^{i}\\
<\mu_{i} \mu_{i}^{T}>= T_{\mu}^{i^{-1}} + m_{\mu}^{i}m_{\mu}^{i^{T}}\\
<\lambda_{n}^{i}>=[a^{i}_{n1} ,,  a^{i}_{nd} ]\\
<\lambda_{n}^{i} \lambda_{n}^{i^{T}}>= \sum_{j=1}^{d}(a_{nj}^{i^{2}} +b_{nj}^{i})\\
<T_{i}>= \nu _{T}^{i} V_{T}^{i^{-1}}\\
<ln|T_{i}|>=\sum_{s=1}^{d}\Psi(\frac{\nu_{T}^{i} + 1 - s}{2}) +dln2 - ln|V_{T}^{i}| \\ 
\end{split}
\end{equation}

\textbf{calculating the Elbo function---------------------------------------------------------------------------------}\\

\begin{equation}
\begin{split}
Elbo = <lnP(D|\lambda)> + <lnP(\lambda|\mu , T , S))> + <lnP(S)> + <lnP(\mu)>+ <lnP(T)>\\ - <lnQ(S)> - <lnQ(\lambda)> - <lnQ(\mu)> - <lnQ(T)>
\end{split}
\end{equation}

\begin{equation}
<lnP(D|\lambda)> =   \sum_{i=1}^{M}\sum_{n=1}^{N}<S_{in}>(\sum_{j=1}^{d}(-<e^{\lambda_{nj}^{i}}>+ <\lambda_{nj}^{i}>x_{nj}  - logx_{nj}! )
\end{equation}

\begin{equation}
\begin{split}
 <lnP(\lambda|\mu , T , S)> = \sum_{i=1}^{M}\sum_{n=1}^{N}<S_{in}> (1/2<ln|T_{i}|> - d/2ln(2\pi)\\
- 1/2Tr(<T_{i}> ( <\lambda_{n}^{i}\lambda_{n}^{i^{T}}>-<\lambda_{n}^{i}><\mu^{i}>^{T} - <\mu^{i}><\lambda_{n}^{i}>^{T}+<\mu^{i}\mu^{i^{T}}>))))
\end{split}
\end{equation}

\begin{equation}
     <lnP(S)> = 2\frac{Md}{2}ln(\frac{\beta }{2\pi}) - \frac{\beta}{2}\sum_{i=1}^{M}<\mu_{i}^{T}\mu_{i}>
\end{equation}

\begin{equation}
    <ln P(\mu)> = 2\frac{Md}{2}ln\frac{\beta }{2\pi} - \frac{\beta }{2}\sum_{i=1}^{M}<\mu_{i}^{t}\mu_{i}>
\end{equation}
\begin{equation}
\begin{split}
 <lnP(T)> =  M \left \{  -\frac{\nu d}{2}ln2 - \frac{d(d-1)}{4}ln\pi -\sum_{s=1}^{d}ln\Gamma (\frac{\nu + 1 - s}{2}) + (\nu)/2\ln|V|\right \} \\ + \frac{\nu - d -1}{2}\sum_{i=1}^{M}<ln|T_{i}|> - 1/2Tr(V\sum_{i=1}^{M}<T_{i}>)
\end{split}
\end{equation}

\begin{equation}
    <lnQ(S)>= \sum_{i=1}^{M}\sum_{n=1}^{N}<S_{in}>ln<S_{in}>
\end{equation}
 
\begin{equation}
    <lnQ(\lambda)> = -\sum_{i=1}^{M}\sum_{n=1}^{N}\sum_{j=1}^{d} 1/2ln(b^{i}_{nj})
\end{equation}

\begin{equation}
    <lnQ(\mu)> = \sum_{i=1}^{M}\left \{ -d/2(1+ln2\pi) + 1/2ln|T^{i}_{\mu}|\right \} 
\end{equation}

\begin{equation}
    \begin{split}
        <lnQ(T)>= \sum_{i=1}^{M}( -\frac{\nu_{T}^{i}d}{2}ln2 -\frac{d(d-1)}{4}ln\pi -\sum_{i=1}^{d}ln\Gamma (\frac{\nu_{T}^{i} + 1 -s}{2}) \\+\nu_{T}^{i}/2ln|V_{T}^{i}| +\frac{\nu_{T}^{i} - d -1}{2}<ln|T_{i}|> -1/2Tr(V_{T}^{i}<T_{i}>) ) 
    \end{split}
\end{equation}
 
\textbf{E-step to optimize the elbo function based on $\lambda^{i}_{nj}$--------------------------------------------------------}
\begin{equation}
\begin{split}
Elbo =  \sum_{i=1}^{M} \sum_{n=1}^{N}P_{in}\sum_{j=1}^{d}\left \{ -<e^{\lambda_{nj}^{i}}> + <\lambda_{nj}^{i}>x_{nj} -1/2t_{jj}^{i}<(\lambda_{nj}^{i} - \mu_{j}^{i})^2>\right \} \\ - \sum_{i=1}^{M} \sum_{n=1}^{N}P_{in}\sum_{j=1}^{d}\sum_{k=1, \neq j}^{d}\left \{ t_{jk}^{i}<(\lambda_{nj}^{i} - \mu_{j}^{i})(\lambda_{nk}^{i} - \mu_{k}^{i})>\right \}  -\sum_{i=1}^{M}\sum_{n=1}^{N}\sum_{j=1}^{d} 1/2ln(b^{i}_{nj}) + constant
\end{split}    
\end{equation}

\begin{equation}
\begin{split}
Elbo =  \sum_{i=1}^{M} \sum_{n=1}^{N}P_{in}\sum_{j=1}^{d}\left \{ -e^{(a_{nj}^{i} + b_{nj}^{i}/2)} + a_{nj}^{i}x_{nj} -1/2<t_{jj}>^{i}(a_{nj}^{i^{2}} + b_{nj}^{i}+ m_{\mu_{j}}^{i^{2}} -2a_{nj}^{i}m_{\mu_{j}}^{i})\right \} \\- \sum_{i=1}^{M} \sum_{n=1}^{N}P_{in}\sum_{j=1}^{d}\sum_{k=1, \neq j}^{d}\left \{ <t_{jk}>^{i}(a_{nj}^{i}a_{nk}^{i} - a_{nj}^{i}m_{\mu_{k}}^{i} - a_{nk}^{i}m_{\mu_{j}}^{i} + <\mu_{j}^{i}\mu_{k}^{i}> )\right \}  \\ -\sum_{i=1}^{M}\sum_{n=1}^{N}\sum_{j=1}^{d} 1/2ln(b^{i}_{nj}) + constant
\end{split}    
\end{equation}
\\
To optimize the Elbo function regarding $a$ and $b$, we find the partial derivative of the Elbo based on a and b. and seek the roots for the new equations. \\
\begin{equation}
    \begin{split}
        \frac{\partial Elbo}{\partial a_{nj}^{i}} = 0 =  Ae^{a_{nj}^{i}} + Ba_{nj}^{i} + C\\
A = e^{b_{nj}^{i}/2}\\
B = (\nu_{T}^{i}V_{T}^{i^{-1}})_{jj}\\
C = -(\nu_{T}^{i}V_{T}^{i^{-1}})_{jj}m_{\mu_{j}}^{i}- x_{nj} +\sum_{k=1, \neq j}^{d}(\nu_{T}^{i}V_{T}^{i^{-1}})_{jk}(m_{\mu_{k}}^{i} -a_{nk}^{i})
    \end{split}
\end{equation}

\begin{equation}
\frac{\partial Elbo}{\partial b_{nj}^{i}} = 0 =  P_{in}e^{a_{nj}^{i}}e^{b_{nj}^{i}/2} + P_{in}(\nu_{T}^{i}V_{T}^{i^{-1}})_{jj} +\frac{1}{b_{nj}^{i}}
\end{equation}

\textbf{M-step to update the mixing coefficients---------------------------------------------------------------}\\
In this step the mixing coefficient for each component is updated based on the following equation:\\

$\pi_{i} = 1/N\sum_{n=1}^{N}p_{in}$

\subsection{Performance evaluation and datasets}

Synthetic sample-taxa count matrices were generated in order to assess
the performance of the VI in detection the number of components in mixture of Poisson log-normal distribution compared to BIC method.


{\bf Synthetic data generation:}
Each sample-taxa count matrix $X$ was produced
by combining samples generated from $K$ component MPLN distributions,
where component $l$ generated $n_l$ samples ($d$-length count vectors), and
such that $\pi_{l}=\frac{n_l}{n}$ and $n=\Sigma_{l=1}^{K}n_{l}$. For
each component $l$, the $d \times d$ covariance matrix of its MPLN
distribution was derived from a randomly generated $d \times d$
positive definite precision matrix. The mean
vector for each MPLN component was a random $d$-length
vector.  \\
To evaluate our algorithm, we generated sample-taxa count matrices $X$ for the following four sets of parameters
\begin{itemize}
\item ($d=50,n_1=(100,200),K=1$),
\item ($d=50,n_1=n_2=(100,200),K=2$),
\item ($d=50,n_1=n_2=n_3=(100,200),K=3$),
\end{itemize}

In addition, 10 replicates were generated for each parameter combination. In total, 20 datasets were generated for each specific value of $K$.\\
In the preprocessing stage, Principle Component Analysis(PCA) is applied to the normalized data. We selected number of components by setting $90\%$ cut-off threshold on cumulative variance. After PCA another normalization is applied to the data to make the convergence in the optimization process faster. \\
Initial value for $v$ degrees of freedom is selected the same as detected number of components in PCA. larger values of v (for fixed scale matrix V) results in smaller values in Precision Matrix. For scale matrix V as diagonal matrix, larger values on diagonal tend to make Precision Matrix to have smaller entries; and vice versa. Initial matrix for V was selected a $d \times d$ diagonal matrix with diagonal entries equal to 1e-1. For beta, initial small value of 1e-6 is selected.  \\
Table 6.1 shows percent of corrected detection in model selection for different values of $K$. Not only does VI works better than BIC, it is more efficient computationally as only on run of VI algorithm is required to detect the right $K$.\\

\begin{table*}[h]
\caption{Values in this table shows percent of corrected detection in model selection. Not only does VI works better than BIC, it is more efficient computationally as it converges in a single training run.      }
\centering
\begin{tabular}{c c c c c c}
    \toprule
    \midrule
    \multirow{2}[4]{*}{MODEL SELECTION METHOD} & \multicolumn{5}{c}{NUMBER OF COMPONENTS IN DATA }\\ 
    \cmidrule(rl){2-6}
    & K  =  1  & K  =  2 & K  =  3  \\ 
    \cmidrule(r){1-1}\cmidrule(l){2-6}
    \multicolumn{1}{l}{BIC (PREVIOUS METHOD)}& 50$\%$ & 50$\%$ & 25$\%$   \\
    \multicolumn{1}{l}{VI (PROPOSED METHOD)}& 60$\%$ & 80$\%$ & 75$\%$    \\
    \midrule
    \bottomrule
\end{tabular}

\end{table*}

\section{Summary}
In this chapter, we presented a practical and efficient framework to find the number of components ($K$) in modeling a sample-taxa matrices with a mixture model. Previous methods such as BIC are not computationally efficient as they need to run the training dataset on several different models to find the best one, while VI gets the best model in one training run. Furthermore they are not as precise as VI. We have also showed that implementing VI for model selection in tensorflow is not practical as needs the hyper parameters to be tuned for each new dataset. We have shown that by driving a closed form formulation for VI in model selection, we will have a precise, practical and computationally efficient method compared to the previous ones. 
 
\section{Summary}
In this chapter, we presented a practical and efficient framework to find the number of components ($K$) in modeling a sample-taxa matrices with a mixture model. Previous methods such as BIC are not computationally efficient as they need to run the training dataset on several different models to find the best one, while VI gets the best model in one training run. Furthermore they are not as precise as VI. We have also showed that implementing VI for model selection in tensorflow is not practical as needs the hyper parameters to be tuned for each new dataset. We have shown that by driving a closed form formulation for VI in model selection, we will have a precise, practical and computationally efficient method compared to the previous ones.

\chapter{SUMMERY}
This dissertation addressed the scenario when the sample-taxa matrix is associated with $K$ microbial networks and considers the computational problem of inferring $K$ microbial networks from a given sample-taxa matrix. \\
In chapter one, we described the motivation behind defining the mentioned computational problem. A literature review on previous methods to infer one single microbial network from a microbial data was included in the next chapter.\\
In chapter three, we presented MixMPLN, a mixture model framework and network inference algorithms, to analyze sample-taxa matrices that are associated with $K$ microbial networks. \\ 
In chapter four, We introduced MixMCMC to infer multiple networks from microbial count data. It is based on the MM principle applied in a mixture model framework and build off of the MixMPLN framework that we introduced previously\cite{decode2}. We used synthetic datasets to evaluate these approaches with respect to recovering the original graph topologies, for different types of graphs. While the AUC values increase with increasing sample size of the input data, we observe that this is a function of the graph type. The methods tend to be able to recover the topologies for band graphs and cluster graphs more accurately in comparison to topologies for scale-free graphs and random graphs. However, MixMCMC is significantly slower.\\
In chapter five, we introduced MixGGM, to infer multiple networks from microbial count data. We used synthetic datasets to evaluate these approaches with respect to recovering the original graph topologies, for different types of graphs. While the AUC values increase with increasing sample size of the input data, we observe that this is a function of the graph type. The methods tend to be able to recover the topologies for band graphs and cluster graphs more accurately in comparison to topologies for scale-free graphs and random graphs. Another observation is that MixGGM (which uses log-ratio transformation) generally tends to perform better compared to MixMPLN. However, the MPLN distribution, which forms the basis of the MixMPLN framework, allows for the inclusion of biological and/or technical co-variates in the Poisson layer to control for confounding factors \cite{Ma2008 , Biswas}. \\
In the next chapter (chapter 6), we presented a practical and efficient framework to find the number of components ($K$) in modeling a sample-taxa matrices with a mixture model. Previous methods such as BIC are not computationally efficient as they need to run the training dataset on several different models to find the best one, while VI gets the best model in one training run. Furthermore they are not as precise as VI. We have also showed that implementing VI for model selection in tensorflow is not practical as needs the hyper parameters to be tuned for each new dataset. We have shown that by driving a closed form formulation for VI in model selection, we will have a precise, practical and computationally efficient method compared to the previous ones.\\
The contributions of this dissertation include 1) new frameworks to generate synthetic sample-taxa count data; 2)novel methods to combine mixture modeling with probabilistic graphical models to infer multiple interaction/association networks from microbial count data; 3) dealing with the compositionality aspect of microbial count data;4) extensive experiments on real and synthetic data; 5)new methods for model selection to infer correct value of $K$.\\









\backmatter

\end{document}